\documentclass[final]{cvpr}

\usepackage{times}
\usepackage{epsfig}
\usepackage{graphicx}
\usepackage{amsmath}
\usepackage{amssymb}
\usepackage{subcaption}
\usepackage[ruled,vlined]{algorithm2e}
\usepackage{multirow}
\usepackage{amsthm}
\usepackage{amsfonts}
\usepackage{enumitem}
\usepackage{color}
\usepackage{makecell}
\usepackage[utf8]{inputenc}

\usepackage{booktabs}

\DeclareMathOperator*{\argmin}{arg\,min}

\usepackage[pagebackref=true,breaklinks=true,colorlinks,bookmarks=false]{hyperref}

\newtheoremstyle{sltheorem}
{}                
{}                
{\slshape}        
{}                
{\scshape}       
{.}               
{ }               
{}                
\theoremstyle{sltheorem}
\newtheorem{thm}{Theorem}

\begin{document}

\title{ProtoPShare: Prototype Sharing for Interpretable Image Classification and Similarity Discovery}

\author{Dawid Rymarczyk$^{1, 2}$\\
\and
Łukasz Struski$^{1}$
\and
Jacek Tabor$^{1}$
\and
Bartosz Zieliński$^{1, 2}$
\and 
$^1$ Faculty of Mathematics and Computer Science, Jagiellonian University $6$~\L{}ojasiewicza Street, 30-348 Krak\'ow, Poland
\and
$^2$ Ardigen SA $76$~Podole Street, 30-394 Krak\'ow, Poland
}

\maketitle

\begin{abstract}
In this paper, we introduce ProtoPShare, a self-explained method that incorporates the paradigm of prototypical parts to explain its predictions. 
The main novelty of the ProtoPShare is its ability to efficiently share prototypical parts between the classes thanks to our data-dependent merge-pruning. Moreover, the prototypes are more consistent and the model is more robust to image perturbations than the state of the art method ProtoPNet. We verify our findings on two datasets, the CUB-200-2011 and the Stanford Cars.
\end{abstract}


\section{Introduction}

Broad application of deep learning in domains like medical diagnosis and autonomous systems enforces models to explain their decisions. Ergo, more and more methods provide human-understandable justifications for their output~\cite{arya2019one,brendel2019approximating,chen2019looks,chen2020concept,fong2019understanding,fu2017look,guidotti2019black,rebuffi2020there}. Some of them are inspired by the human brain and how it explains its visual judgments by pointing to prototypical features that an object possesses~\cite{salakhutdinov2012one}. I.e., a certain object is a car because it has tires, roof, headlights, and horn.

\begin{figure}[t]
    \centering
    \includegraphics[width=0.43\textwidth]{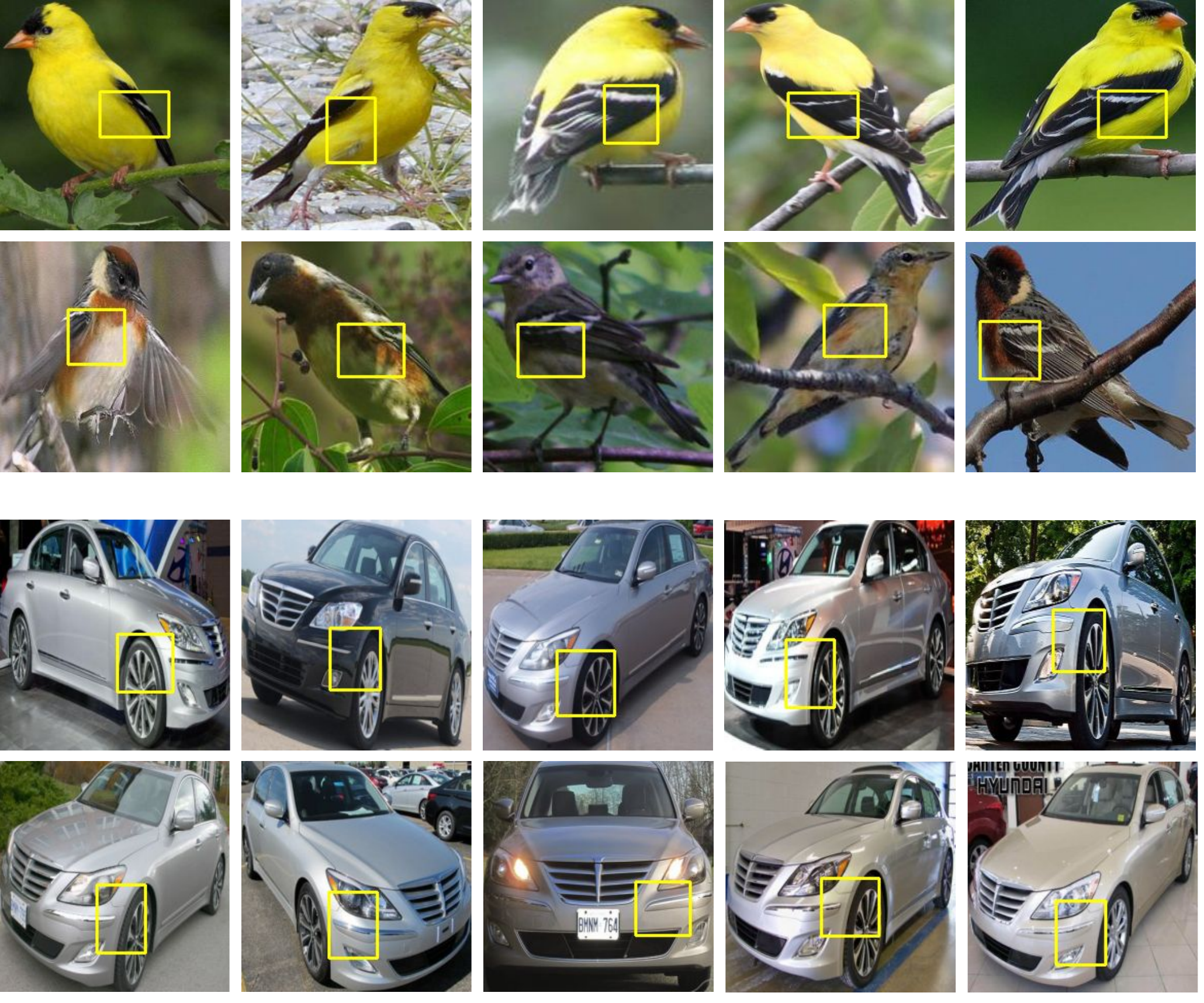}
    \caption{Two pairs of semantically similar prototypes (first representing \textit{bright belly with grayish wings} and second representing \textit{fender}) that are distant in the representation space but close according to our data-dependent similarity. Notice that each prototype is represented by one row of $5$ closest images' parts (marked with yellow bounding-boxes). More examples are presented in the Supplementary Materials.}
    \label{fig:dd_di_similarity}
\end{figure}

Recently introduced \textit{Prototypical Part Network} (\textit{ProtoPNet})~\cite{chen2019looks} applies this paradigm by focusing on parts of an image and comparing them with \textit{prototypical parts} of a given class. This comparison is achieved by pushing parts of an image and prototypical parts through the convolutional layers, obtaining their representation, and computing similarity between them. We will refer to the representations of prototypical parts as \textit{prototypes}.
\begin{figure}[t]
    \hspace{-5pt}
    \includegraphics[width=0.49\textwidth]{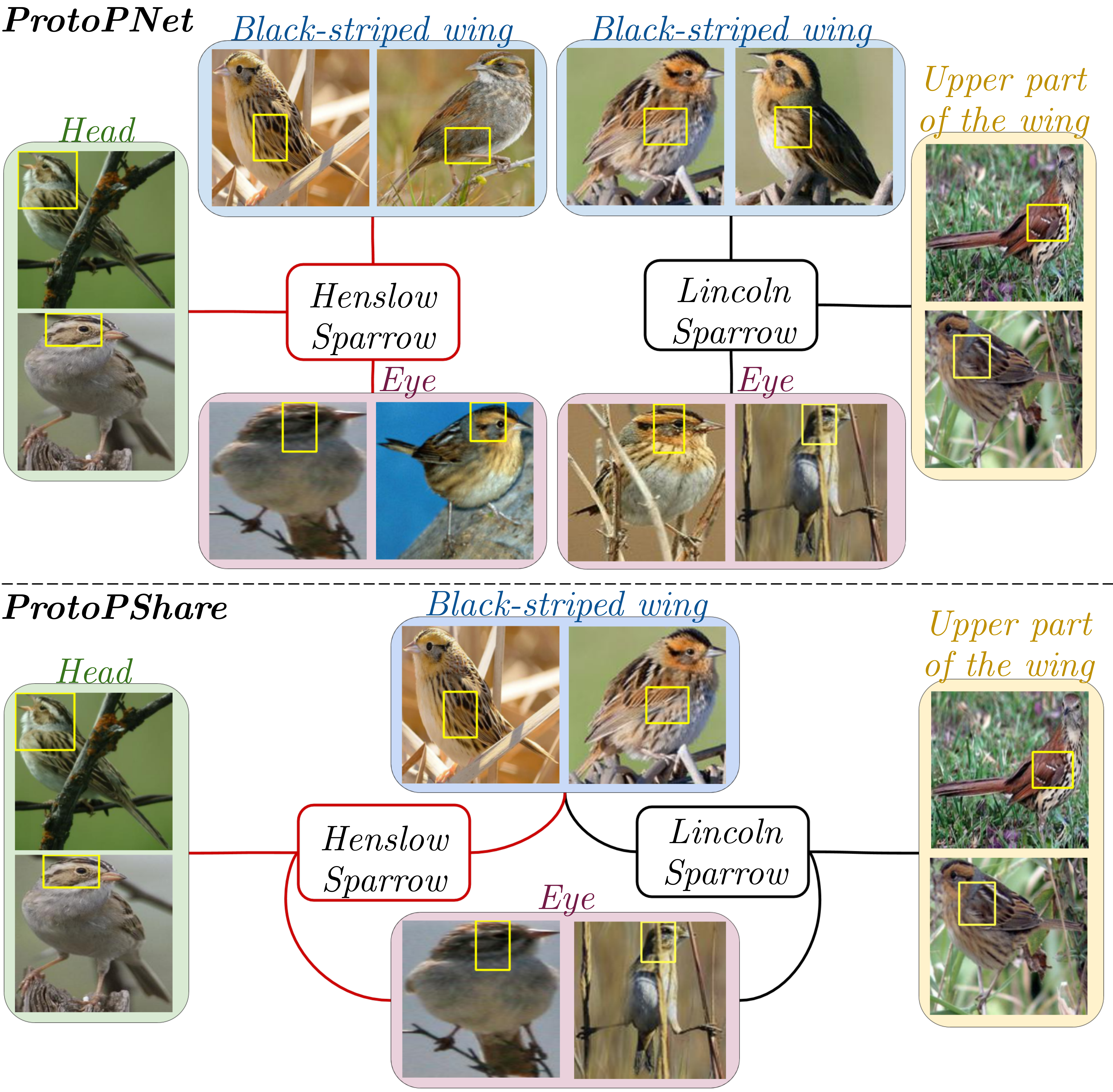}
    \caption{The number of prototypes in ProtoPNet is large because each of them is assigned to only one class. In contrast, our ProtoPShare shares prototypes between classes. E.g., in this example, both \textit{Henslow sparrow} and \textit{Lincoln sparrow} have separate prototypes corresponding to \textit{Eye} in ProtoPNet, which is merged into one shared prototype in ProtoPShare. As a result, we decrease the number of prototypes, which positively influences the model interpretability.}
    \label{fig:graph_supp}
\end{figure}
ProtoPNet is a self-explaining model that generates intuitive explanations and achieves accuracy comparable with its analogous non-interpretable counterparts. However, it has limited applicability due to two main reasons. Firstly, the number of prototypes is large because each of them is assigned to only one class. It negatively influences the interpretability whose much-desirable properties are small size and low complexity~\cite{doshi2017roadmap,wang2019gaining}. Secondly, due to the training that pushes away the prototypes of different classes in the representation space, the prototypes with similar semantic can be distant (see Figure~\ref{fig:dd_di_similarity}). Therefore, the predictions obtained from the network can be unstable.

\begin{figure}[t]
    \centering
    \includegraphics[width=0.47\textwidth]{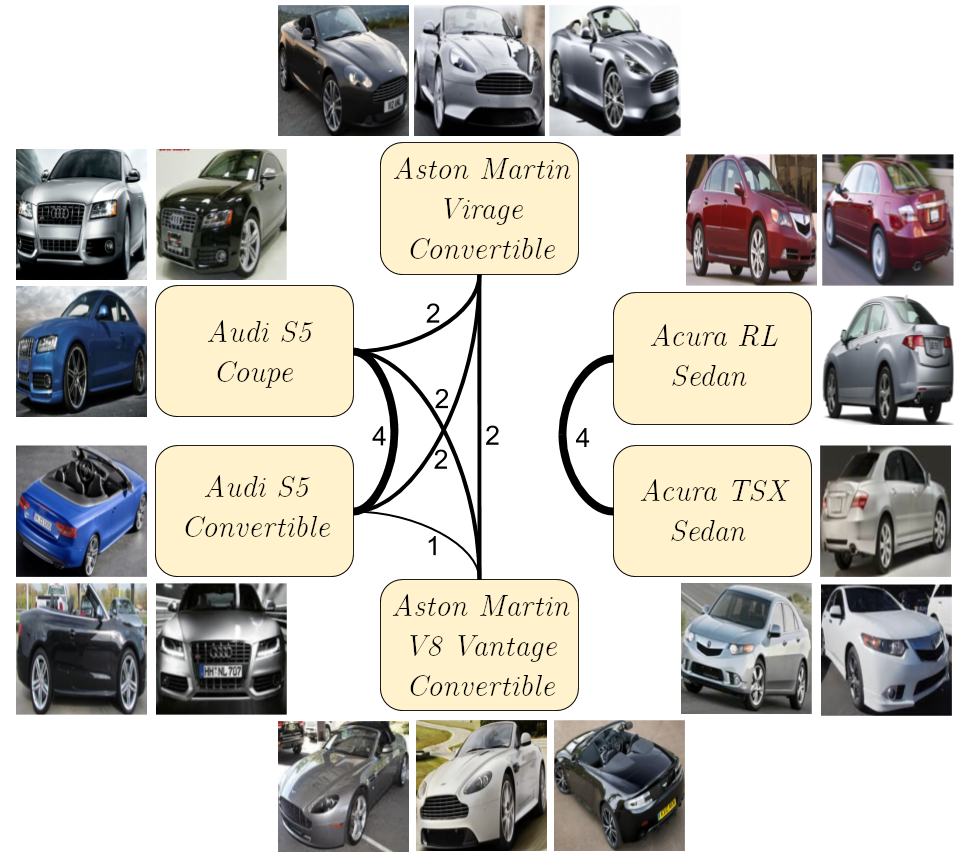}
    \caption{Inter class similarity visualization as a graph generated based on the prototypes shared in ProtoPShare where each node corresponds to a class and the strength of edge between two nodes corresponds to the number of shared prototypes. E.g., \textit{Audi S5 Coupe} shares $4$ prototypes with \textit{Audi S5 Convertible} but does not share prototypes with \textit{Acura RL Sedan}. This graph can be used to find similarities between the classes or to cluster them into groups. Notice that each class is represented by three images located around the node.}
    \label{fig:graph_cars}
\end{figure}

In this paper, we address those limitations by introducing the \textit{Prototypical Part Shared} network (\textit{ProtoPShare})\footnote{The code will be available in the camera ready version.} that shares the prototypes between the classes, as presented in Figure~\ref{fig:graph_supp}. As a result, the number of prototypes is relatively small and prototypes with similar semantic are close to each other. Additionally, it is possible to discover similarities between the classes, as presented in Figure~\ref{fig:graph_cars}.
ProtoPShare method consists of two phases, initial training and prototypes' pruning. In the first phase, the model is trained using exclusive prototypes and a loss function described in~\cite{chen2019looks}. In the second phase, the pruning proceeds in steps that merge defined portion of the most similar prototypes. For this purpose, we introduce the data-dependent similarity described in Section~\ref{sec:method} that finds prototypes with similar semantics, even if they are distant in representation space. We demonstrate the superiority of the ProtoPShare approach by comparing it to the other methods based on the prototypes' paradigm. Thus, the main contributions of the paper are:\vspace{-6pt}
\begin{itemize}
    \item We construct ProtoPShare, a self-explained method built on the paradigm of prototypical parts that shares prototypes between the classes.
    \item We introduce data-dependent similarity that can find semantically similar prototypes even if they are distant in the representation space.
    \item Compared to the state of the art approach ProtoPNet, we reduce the number of prototypes and enable finding the prototypical similarities between the classes.
\end{itemize}
\vspace{-6pt}

This paper is organized as follows. Section~\ref{sec:rw} investigates related works on interpretability and pruning. In Section~\ref{sec:method}, we introduce our ProtoPShare method together with its theoretical understanding. Section~\ref{sec:exp} illustrates experimental results on CUB-200-2011 and Stanford Cars datasets, while Section~\ref{sec:interpretability} concentrates on ProtoPShare interpretability. Finally, we conclude the work in Section~\ref{sec:dc}.

\section{Related works}
\label{sec:rw}

ProtoPShare is a self-explained method with a strong focus on the prototypes' pruning. Therefore, in the related works, we consider the articles about interpretability and pruning.

\begin{figure*}[t]
    \centering
    \includegraphics[width=\textwidth]{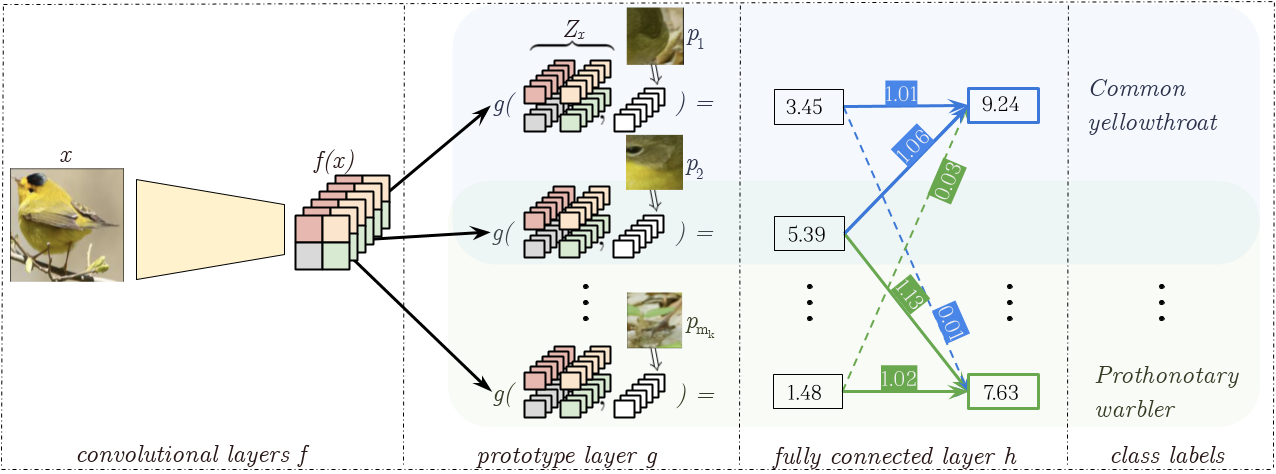}
    \caption{ProtoPShare consists of convolutional layers $f$, followed by a prototype layer $g$, and a fully connected layer $h$. Layer $g$ contains prototypes that are shared between the classes (e.g., in this example, prototype $p_2$ is assigned both to class \textit{Common yellowthroat} and \textit{Prothonotary warbler} which corresponds to stronger connections between $p_2$ and both classes in layer $h$). As a result, we reduce the number of prototypes.}
    \label{fig:protopnet}
\end{figure*}

\paragraph{Interpretability.} Interpretability approaches can be divided into post hoc and self-explaining methods~\cite{arya2019one}. Post hoc techniques include saliency maps, showing how important each pixel of a particular image is for its classification~\cite{rebuffi2020there,selvaraju2017grad,selvaraju2019taking,simonyan2013deep}. Another technique called concept activation vectors provides an interpretation of the internal network state in terms of human-friendly concepts~\cite{chen2020concept,ghorbani2019towards,kim2018interpretability,yeh2019completeness}. Other methods analyze how the network’s response changes for perturbed images~\cite{fong2019understanding,fong2017interpretable,ribeiro2016should}. Post hoc methods are easy to use in practice as they do not require changes in the architecture. However, they can produce unfaithful and fragile explanations~\cite{adebayo2018sanity}.

Therefore, as a remedy, self-explainable models were introduced by applying the attention mechanism~\cite{fu2017look,sundararajan2017axiomatic,xiao2015application,zheng2017learning,zhou2018interpretable} or the bag of local features~\cite{brendel2019approximating}. Other works~\cite{guidotti2019black,li2017deep,puyol2020interpretable} focus on exploiting the latent feature space obtained, e.g., with adversarial autoencoder.

From this paper's perspective, the most interesting self-explaining methods are prototype-based models represented, e.g., by Prototypical Part Network~\cite{chen2019looks} with a hidden layer of prototypes representing the activation patterns. A similar approach for hierarchically organized prototypes is presented in~\cite{hase2019interpretable} to classify objects at every level of a predefined taxonomy.  Some works concentrate on transforming prototypes from the latent space to data space~\cite{hase2019interpretable}. The others try to apply prototypes to other domains like sequence learning~\cite{ming2019interpretable} or time series analysis~\cite{gee2019explaining}.

\paragraph{Pruning.}
Pruning is mostly used to accelerate deep neural networks by removing unnecessary weights or filters~\cite{liu2018rethinking}. The latter can be roughly divided into data-dependent and data-independent filter pruning~\cite{he2019filter}, depending on whether the data is used or not to determine the pruned filters.

The basic type of data-independent pruning removes filters with the smallest sum of absolute weights~\cite{li2016pruning}. More complicated approaches designate irrelevant or redundant filters using scaling parameters of batch normalization layers~\cite{ye2018rethinking} or the concept of geometric median~\cite{he2019filter}.

Data-dependent pruning can be solved as an optimization problem on the statistics computed from its subsequent layer~\cite{luo2017thinet}, by minimizing the reconstruction error on the subsequent feature map using Lasso method~\cite{he2017channel}, or by using a criterion based on Taylor expansion that approximates the change in the cost function induced by pruning~\cite{molchanov2019pruning}. Filters' importance can also be propagated from the final response layer to minimize its reconstruction error~\cite{yu2018nisp}. Another possibility is to introduce additional discrimination-aware losses and select the most discriminative channels for each layer~\cite{zhuang2018discrimination}. Finally, feature maps can be clustered and replaced by the average representative from each cluster~\cite{wang2018exploring}. Moreover, filters can be pruned together with other structures of the network~\cite{lin2019towards}.

Our ProtoPShare extends the ProtoPNet~\cite{chen2019looks} by the shared prototypes obtained with the data-dependent pruning based on the feature maps.

\section{ProtoPShare}
\label{sec:method}

In this section, we first describe the architecture of ProtoPShare and then define our data-dependent merge-pruning algorithm. Finally, we provide a theoretical understanding of how the prototypes' merge affects classification accuracy.

\paragraph{Architecture.}
The architecture of ProtoPShare, shown in Figure~\ref{fig:protopnet}, consists of convolutional layers $f$, followed by a prototype layer $g$, and a fully connected layer $h$ (with weight $w_h$ and no bias). Given an input image $x \in X$, the convolutional layers extract image representation $f(x)$ of shape $H\times W\times D$. For the clarity of description, let $Z_x=\{z_i\in f(x) : z_i\in \mathbb{R}^D, i=1..HW\}$. Then, for each class $k$, the network learns $m_k$ prototypes $P_k= \{p_i\}_{i=1}^{m_k}$, where $p_i\in \mathbb{R}^D$ represents prototypical parts trained for class $k$. Given a convolutional output $Z_x$ and prototype $p$, ProtoPNet computes the distances between $p$ and $Z_x$ patches, inverts them to obtain the similarity scores, and takes the maximum:
\begin{equation}
    g(Z_x, p) = \max\limits_{z\in Z_x} \log\left(\dfrac{\|z - p\|^2 + 1}{\|z - p\|^2 + \varepsilon}\right)\; \text{ for }\; \varepsilon > 0.
    \label{eq:g_p}
\end{equation}
Finally, the similarity scores produced by the prototype layer ($m_k$ values per class) are multiplied by the weight matrix $w_h$ in the fully connected layer $h$. It produces the output logits that are normalized using softmax to obtain a prediction.

After training with exclusive prototypes and a loss function described in~\cite{chen2019looks}, most representations of the image's parts are clustered around semantically similar prototypes of their true classes, and the prototypes from different classes are well-separated. As a result, the prototypes with similar semantics can be distant in representation space, resulting in unstable predictions. In the next paragraph, we present how to overcome this issue with our data-dependent merge-pruning.

\paragraph{Data-dependent merge-pruning.}
Here, we describe the pruning phase of our method. As the input, it obtains the network trained with exclusive prototypes, and as the output, it returns the network with a smaller number of shared prototypes.

\begin{figure}[t]
    \centering
    \includegraphics[width=0.40\textwidth]{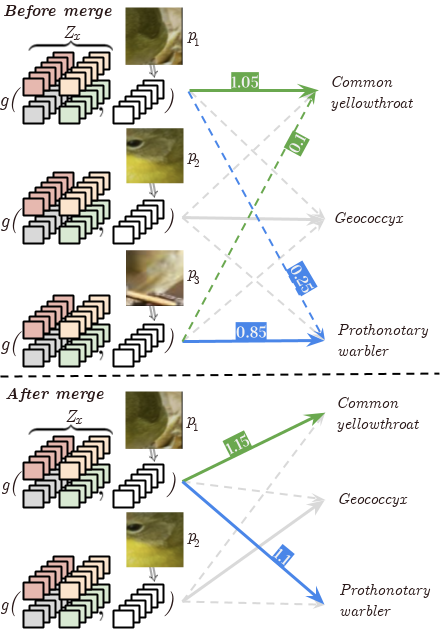}
    \caption{To visualize the process of merging two prototypes, let us assume that prototypes $p_1$, $p_2$, and $p_3$ are assigned (have strong connections) to \textit{Common yellowthroat}, \textit{Geococcyx}, and \textit{Prothonotary warbler}, respectively. Let as also assume that $p_1$ and $p_3$ are similar (e.g., in this example they both present a \textit{birds' leg}). Then, we merge $p_1$ and $p_3$ by summing weights represented as the blue and green arrows, and removing $p_3$ from the network. As a result, $p_1$ has strong connections to both classes to which it is assigned after merge.}
    \label{fig:pruned_sup}
\end{figure}

Let $P$ be the set of all prototypes (after training phase), $K$ be the number of classes, and $\zeta$ be the percentage of prototypes to merge per pruning step. Each step begins with computing the similarities between the pairs of prototypes, then for each pair $(p, \tilde{p})$ among $\zeta$ percent of the most similar pairs, the prototype $p$ is removed together with its weights $w_h(p)$. In exchange, the class to which $p$ was assigned reuses prototype $\tilde{p}$ whose weights $w_h(\tilde{p})$ are aggregated with $w_h(p)$. We present this procedure in Figure~\ref{fig:pruned_sup}. Note that we merge around $\zeta$ (not exactly $\zeta$) percent of prototypes because similarity can be the same for many pairs.

To overcome the problem with prototypes of similar semantic that are far from each other, we introduce a special type of data-dependent similarity.  It uses training set $X$ to generate the representation of all training patches $\bigcup\limits_{x\in X}Z_x$, and then calculates the difference between $g(Z_x, p)$ and $g(Z_x, \tilde{p})$. As a result, the \textit{data-dependent similarity} for pair of prototypes $(p, \tilde{p})\in P^2$ is defined as:
\begin{equation}
d_{DD}(p, \tilde{p}) = \frac{1}{\sum_{x \in X} \left( g(Z_x, p) - g(Z_x, \tilde{p}) \right)^2}.
\label{eq:dDD}
\end{equation}

\paragraph{Theoretical results.}
The following theorem provides some theoretical understanding of how the prototypes' merge affects the classification accuracy. Intuitively, the theorem states that if the pruning merges similar prototypes, then the predictions after merge do not change if predictions before the merge had certain confidence. Due to the page limits, the proof is in the Supplementary Materials.

\begin{thm}
Let $x\in X$ be an input image correctly classified by ProtoPShare $h\circ g_p\circ f$ as class $c$ before the prototypes' merge, and let: \vspace{-5pt}
\begin{enumerate}[label=(A\arabic*)]

    \item some of the class $k$ prototypes remain unchanged $P'_k\subset P_k$ while the others merge into the other classes' prototypes $S_k\subset\bigcup\limits_{i\neq k}P'_i$,
    
    \item $p\in P_k\setminus P'_k$ is merged into $\tilde{p}\in S_k$,

    \item $z_p=\argmin\limits_{z\in \bigcup\limits_{x\in X}Z_x} \|z - p\|$ is the representation of the training patch nearest to any prototype $p$, \label{assumption_nearest}

	\item there exist $\delta\in(0, 1)$ such that:
	\begin{enumerate}[label=\alph*), ref=(A\arabic{enumi}\alph*)]
        \setlength{\itemsep}{0pt}
        \setlength{\parskip}{-2pt}
        \setlength{\parsep}{-1pt}

    	\item for $k\neq c$, $p\in P_k\setminus P_k'$ and $\tilde{p}\in S_k$, we suppose that $\|p - \tilde{p}\| \leq\theta\|z_p - p\| - \sqrt{\varepsilon}$ and $\theta = \min(\sqrt{1 + \delta} - 1, 1 - \frac{1}{\sqrt{2-\delta}})$ ($\varepsilon$ comes from function $g$ defined in~\eqref{eq:g_p}), \label{assumption_ineq_a}

    	\item for $p\in P_c\setminus P'_c$ and $\tilde{p}\in S_c$, we suppose that $\|\tilde{p} - p\| \leq(\sqrt{1+\delta}-1)\|z_p - p\|$ and $\|z_p - p\|\leq\sqrt{1-\delta}$,
	\label{assumption_ineq_b}
	\end{enumerate}
    \vspace{-5pt}
	\item for each class $k$, weights connecting class with assigned prototypes equal $1$, and the other weights equal 0 (i.e., $w_h(p) = 1$ for $p\in P_k\cup S_k$ and $w_h(p) = 0$ for $p\in\bigcup\limits_{i\neq k}P_i\setminus S_k$). \label{assumption_weight}
\end{enumerate}
\vspace{-5pt}
\noindent Then after the prototypes' merge, the output logit for class $c$ can decrease at most by $\Delta^c_{\max}$, and the output logit for the other classes $k\neq c$ can increase at most by $\Delta^k_{\max}$, where:
\[
\Delta^k_{\max}:=|P_k\setminus P'_k|\log \left((1+\delta)(2-\delta)\right) \; \text{ for }\; k\in\{1, \ldots, K\}. \]
If the output logits between the top-2 classes are at least $\Delta^c_{\max} + \max\limits_{k\neq c}\big(\Delta^k_{\max}\big)$ apart, then the merge of prototypes does not change the prediction of $x$.
\end{thm}

\section{Experiments}
\label{sec:exp}

\begin{figure}[t]
    \centering
    \includegraphics[width=0.5\textwidth]{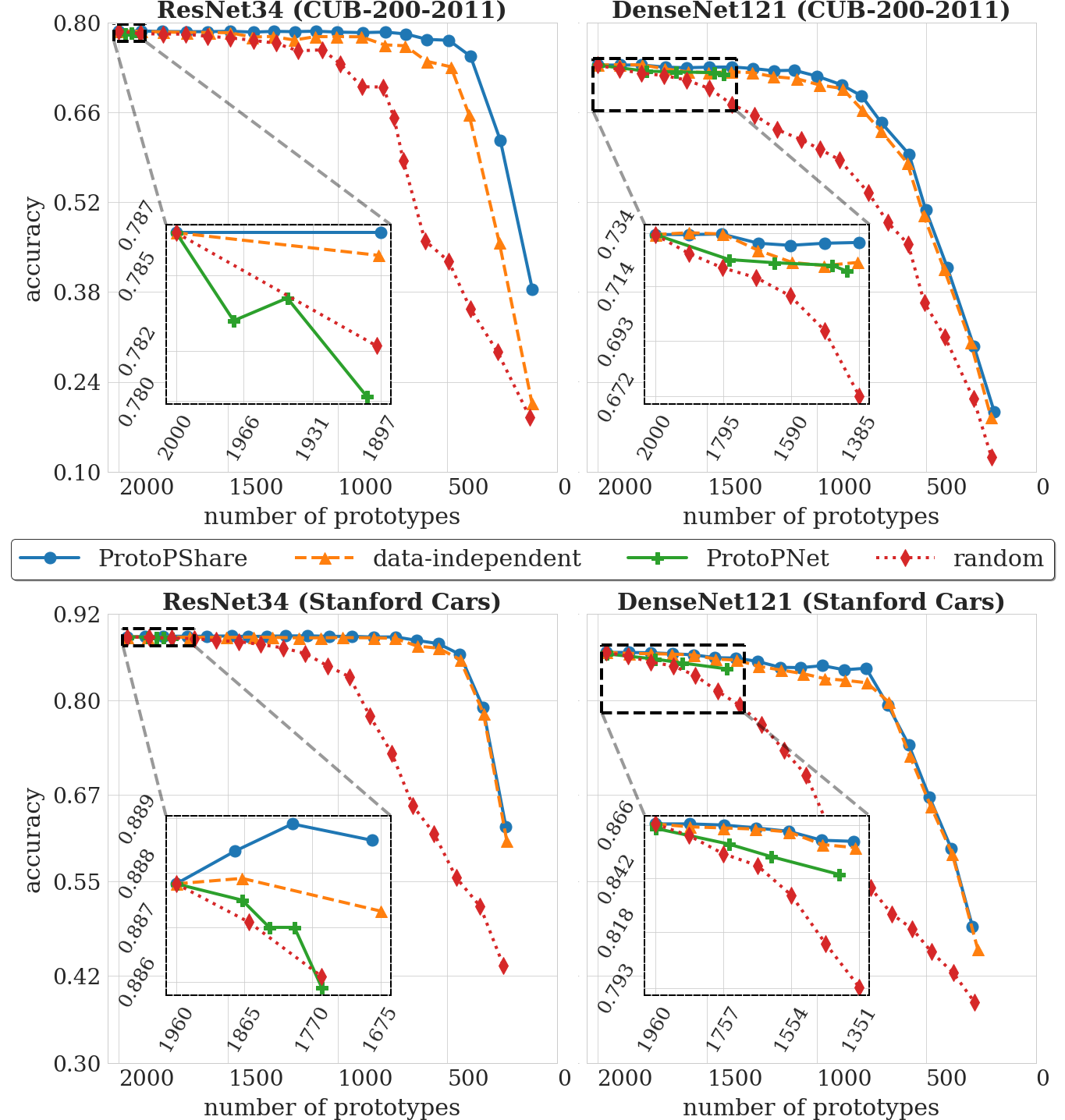}
    \caption{We compare the accuracy (higher is better) of ProtoPShare, ProtoPNet, and variations of our method (data-independent and random) for various pruning rates on CUB-200-2011 and Stanford Cars datasets (first and second row, respectively). ProtoPShare achieves higher accuracy for all pruning rates and works reasonably well even for only $20\%$ of the initial prototypes. Notice that the accuracies for the initial pruning steps are zoomed to enable comparison with ProtoPNet that can prune at most $30\%$ of prototypes. The remaining architectures' results are presented in the Supplementary Materials.}
    \label{fig:prune_diff}
\end{figure}

In this section, we analyze the accuracy and robustness of ProtoPShare in various scenarios of the merge-pruning, and we explain our architectural choices by comparing it to other methods.

We train ProtoPShare to classify  200 bird species from the CUB-200-2011 dataset~\cite{wah2011caltech} and 196 car models from the Stanford Cars dataset~\cite{Krause2013}. As the convolutional layers $f$, we take the convolutional blocks of ResNet-34, ResNet-152~\cite{he2016deep}, DenseNet-121, and DenseNet-161~\cite{huang2017densely} pretrained on ImageNet~\cite{deng2009imagenet}. We set the number of prototypes per class to $10$, which results in $2000$ prototypes for birds and $1960$ prototypes for cars after the first phase of ProtoPShare. Moreover, in the second phase, we assume finetuning with $25$ iterations after each pruning step.

\paragraph{How many prototypes are required?}
In Figure~\ref{fig:prune_diff}, we compare the accuracy of ProtoPShare with ProtoPNet and variations of our method (data-independent and random) for various pruning rates. Data-independent uses inverse Euclidean norm as the similarity measure instead of similarity defined in~\eqref{eq:dDD}, while random corresponds to random joining. The results for ProtoPShare and its variations were obtained from the successive steps of pruning, while the results for ProtoPNet were obtained for different value of $\tau=1..5$ from Appendix S8 in~\cite{chen2019looks}.

One can observe that ProtoPShare achieves higher accuracy for all pruning rates and works reasonably well even for only $20\%$ of the initial prototypes in case of ResNet34. Moreover, data-independent obtains similarly good results at the initial steps of prunings, but its accuracy drops more rapidly with a higher pruning rate. At the same time, ProtoPNet results are between data-independent and random, which works surprisingly well, even for $50\%$ removed prototypes. One should also notice that the ProtoPNet can prune at most $30\%$ of prototypes (e.g., around $200$ prototypes for ResNet34), even for higher values of $\kappa$ and $\tau$ from Appendix S8 in~\cite{chen2019looks}. Finally, for each model, we observe a critical step of pruning with a significant decrease in accuracy. Hence, we suggest to monitor the accuracy of the validation set when applying ProtoPShare in specific domains.

\begin{figure}[t]
    \centering
    \includegraphics[width=0.5\textwidth]{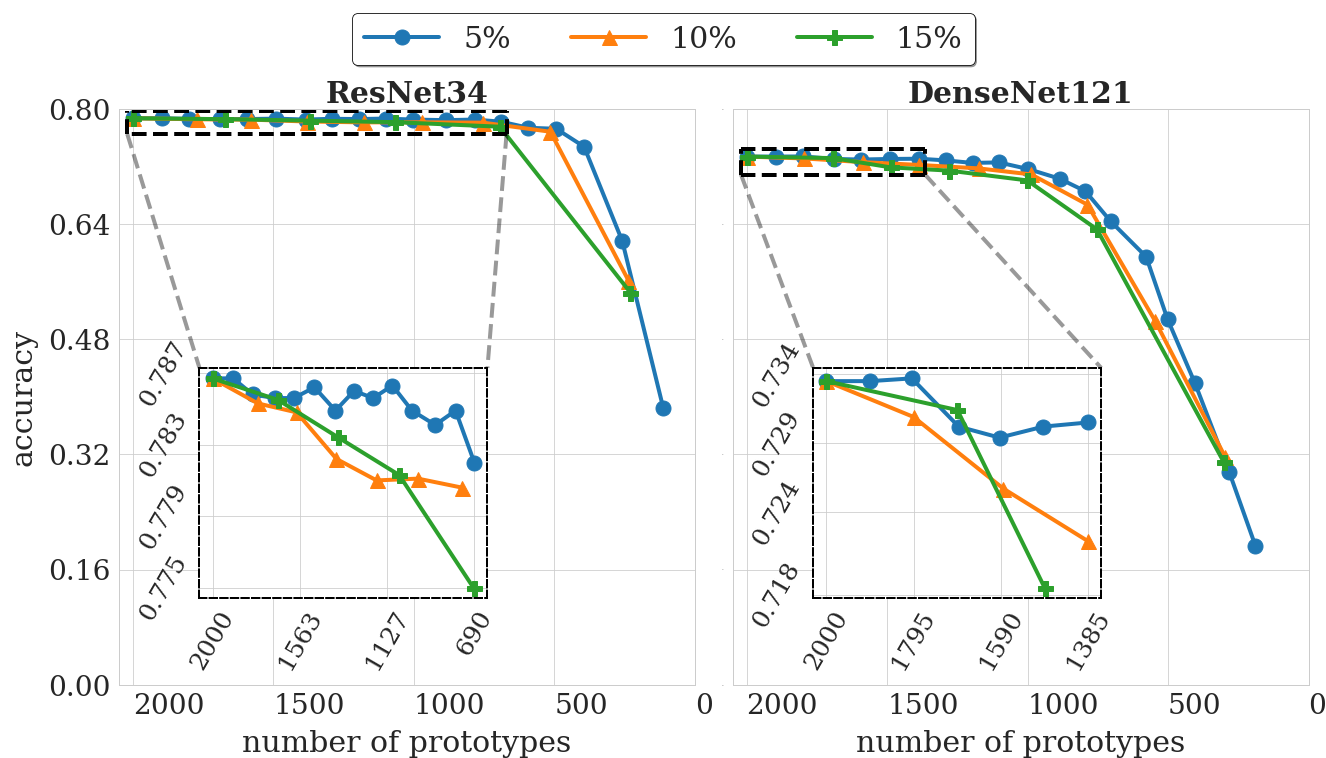}
    \caption{We compare the accuracy (higher is better) of ProtoPShare for different percentages of prototypes merged per pruning step ($5\%$, $10\%$, and $15\%$) on the CUB-200-2011 dataset. We observe that the drop in accuracy is lower for a smaller number of prototypes merged per step. It is expected because small modifications in the set of prototypes make finetuning more effective than in the case of broader changes. Notice that the accuracy for the initial pruning steps is zoomed. The remaining architectures' results are presented in the Supplementary Materials.}
    \label{fig:percents}
\end{figure}

\paragraph{Accuracy vs. step size.}
In Figure~\ref{fig:percents}, we present how the ProtoPShare behaves for different percentages of prototypes merged per pruning step ($\zeta=5\%$, $10\%$, or $15\%$). One can observe that the accuracy is higher for smaller $\zeta$. It is expected because small modifications in the set of prototypes make finetuning with $25$ iterations more effective than in the case of broader changes. Notice that we have not tested $\zeta<5\%$, but we hypothesize that it would further increase the accuracy by extending the computations.

\begin{table}[t]\scriptsize
\setlength{\tabcolsep}{4pt}
\newlength\wexp
\settowidth{\wexp}{Exponential}
\newcolumntype{C}{>{\centering\arraybackslash}p{\dimexpr.5\wexp-\tabcolsep}}

\centering
\begin{tabular*}{\columnwidth}{l@{\;\;}CCCCc@{\;\;}c@{\;\;}c@{\;\;}c@{}}
\toprule
\multirow{3}{*}{Model} & $|P|$ &  \multicolumn{3}{c}{Finetuning after} & \makecell{RN34} & \makecell{RN152} & \makecell{DN121} & \makecell{DN161} \\
 & before & \multicolumn{3}{c}{\makecell{pruning step}} & \multicolumn{4}{c}{$|P|$ in the final model} \\
 & pruning & $f$ & $P$ & $h$ & 400 & 1000 & 1000 & 600 \\
\midrule
ProtoPNet & & \multicolumn{3}{c}{no pruning} & 0.7259 & 0.7151 & 0.4621 & 0.7075 \\
shared in training & & \multicolumn{3}{c}{no pruning} & 0.6833 & 0.6938 & 0.3764 & 0.6823 \\
\midrule
\multirow{3}{*}{\makecell{ProtoPShare\\(ours)}} & 2000 & \checkmark & \checkmark & \checkmark & 0.6355 & 0.6821 & 0.6612 & 0.6526 \\
 & 2000 & & \checkmark & \checkmark & 0.6591 & 0.7161 & 0.6836 & 0.7047 \\
 & 2000 & & & \checkmark & \textbf{0.7472} & \textbf{0.7361} & \textbf{0.7472} & \textbf{0.7645} \\
\toprule
\end{tabular*}
\caption{We present the accuracy (higher is better) for different models with the same final number of prototypes $|P|$ trained on the CUB-200-2011 dataset. E.g., in the case of ResNet34 (RN34), we compare ProtoPShare (first trained with $2000$ prototypes and then pruned to $400$) with its no-pruning version that shares $400$ prototypes between the classes during the training phase, and with the ProtoPNet also trained with $400$ prototypes ($2$ per class). Additionally, we test different finetuning strategies used after ProtoPShare pruning steps. Those strategies include optimizing only the last layer ($h$), optimizing the prototypes and the last layers ($P$ and $h$), and optimizing all the layers ($f$, $P$, and $h$). We conclude that ProtoPShare with the last layer finetuned after pruning step achieves the highest accuracy. Note that RN and DN correspond to ResNet and DenseNet architectures, respectively.}
\label{tab:diff_optims}
\end{table}

\paragraph{Why two training phases?}
To demonstrate the need for two phases in ProtoPShare, we compare it to its no-pruning version that shares prototypes between the classes during the training phase. To make this new version comparable with ProtoPShare, we set the number of prototypes to a small number (e.g., $400$ for ResNet34) and train the model using standard cross-entropy loss. The comparison between this model and ProtoPShare first trained for a large number of prototypes ($2000$) and then pruned to a small number of prototypes (e.g., $400$ for ResNet34) is presented in Table~\ref{tab:diff_optims}. One can observe that ProtoPShare obtains much better results, even doubling its no-pruning version in the case of DenseNet121. Moreover, the no-pruning version does not clearly explain a prediction because all prototypes are assigned to all classes.
In that case, the question arises: does ProtoPNet with a small number of prototypes (e.g., 400 for ResNet34) work
better than above ProtoPShare? The results presented in Table~\ref{tab:diff_optims} again show the superiority of ProtoPShare. It is expected because the latter shares the prototypes between the classes. Finally, Table~\ref{tab:diff_optims} provides the comparison of finetuning strategies that can be used after the pruning step, which clearly shows that only the fully connected layer should be finetuned.

\begin{figure}[t]
    \centering
    \includegraphics[width=0.45\textwidth]{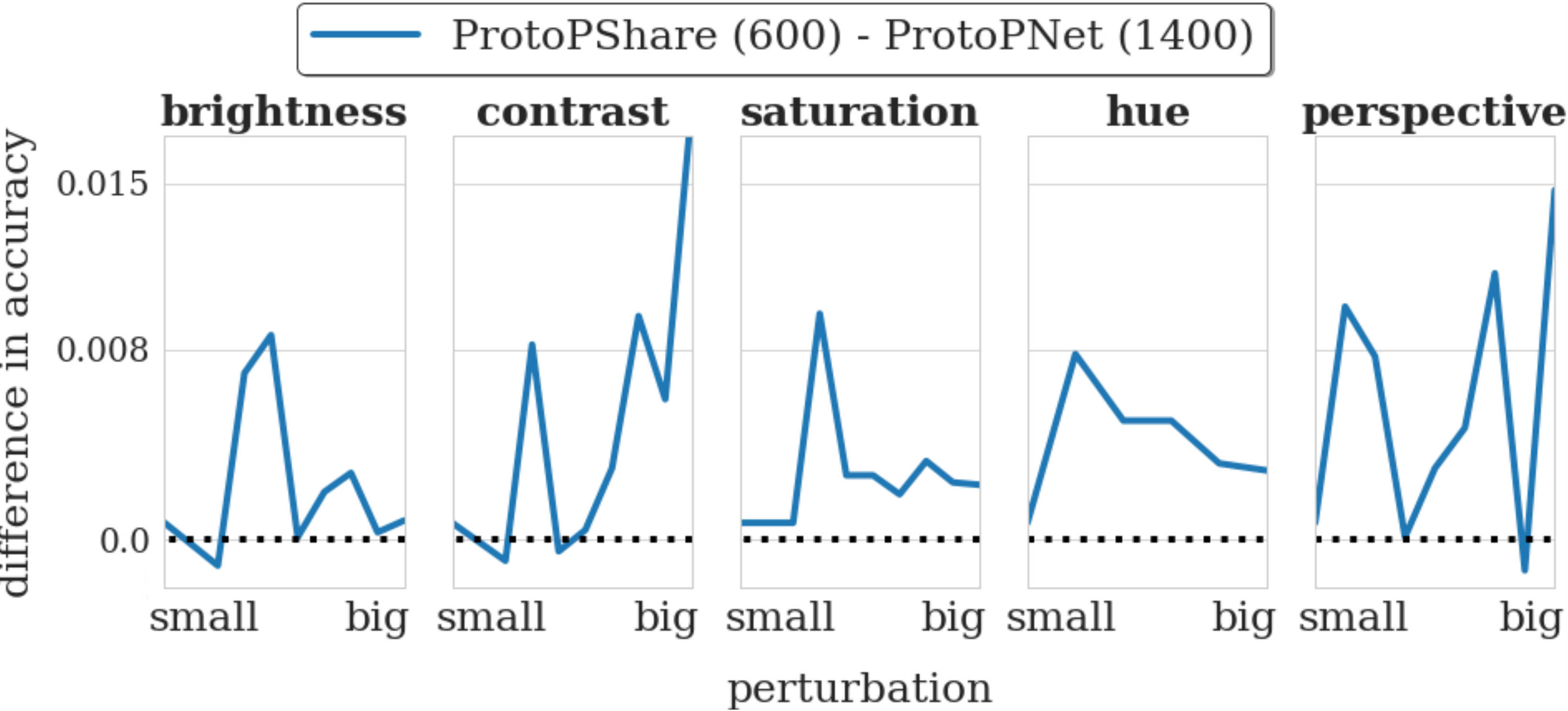}
    \caption{The difference in the accuracy between ProtoPShare with $600$ prototypes and ProtoPNet with $1400$ prototypes (using DenseNet161 architecture) for various image perturbations (positive values are in favor of ProtoPShare). We observe that ProtoPShare is more robust to perturbations in brightness, contrast, saturation, hue, and perspective what suggests that a smaller number of prototypes do not increase the model's susceptibility to image perturbations.}
    \label{fig:perturbations}
\end{figure}

\begin{figure*}[t]
    \centering
    \includegraphics[width=\textwidth]{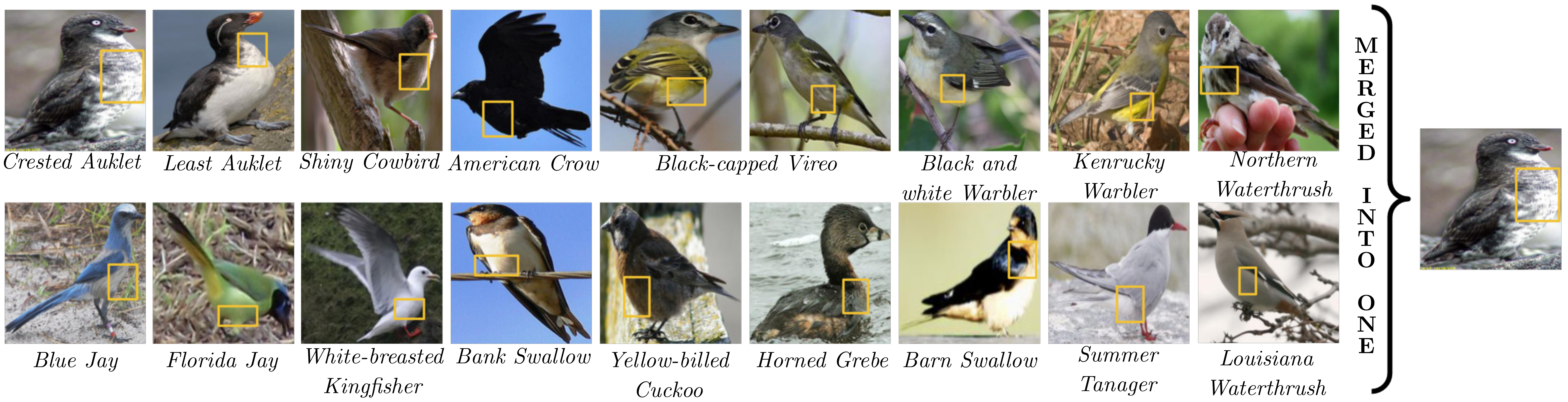}
    \caption{Eighteen prototypes from seventeen classes merged into one prototype by our ProtoPShare method. One can see that all prototypes correspond to the prototypical part of \textit{a belly brighter than the wings}.}
    \label{fig:pruned_sup2}
\end{figure*}

\paragraph{Resistance to perturbations.}
In Figure~\ref{fig:perturbations} we present the difference in accuracy between ProtoPShare trained for $2000$ and then pruned to $600$ prototypes with ProtoPNet trained for $2000$ and then pruned to $1400$ prototypes (using $\kappa=6$ and $\tau=5$ from Appendix S8 in~\cite{chen2019looks}) for various image perturbations. We incorporate different perturbations and magnitudes that modify brightness, contrast, saturation, hue, and perspective using \textit{Torchvision ColorJitter} and \textit{Perspective} transformations with probability of perturbation equal $1$ and perturbation values from range $[0;0.5]$ for hue and $[0;1]$ for the others. As is shown, using the stronger perturbation, the ProtoPShare with $600$ prototypes performs slightly better than ProtoPNet with $1600$ prototypes. Therefore, we conclude that a smaller number of prototypes do not increase the model's susceptibility to image perturbations.

\paragraph{Discussion.}
We confirmed that ProtoPShare achieves better accuracy than other methods for almost all pruning rates. We demonstrated the need for small-steps pruning with finetuning only the last fully connected layer instead of the broader optimization. Additionally, we explained the need for two phases (training and pruning) and verified that the smaller number of prototypes does not negatively influence the model's robustness.

\section{Interpretability of ProtoPShare}
\label{sec:interpretability}

In the section, we focus on the interpretability of ProtoPShare using qualitative results and user study. We explain why the merge of prototypes is better than the original pruning from~\cite{chen2019looks}, illustrate how our model can discover the inter-class similarity, and demonstrate the superiority of data-dependent similarity over its data-independent counterpart.

\paragraph{Why merge-pruning instead of pruning?}
We decided to introduce our merge-pruning algorithm after a detailed analysis of the prototypes pruned by ProtoPNet and concluding that around $30\%$ of them represent significant prototypical parts instead of a background (examples are provided in the Supplementary Materials). It negatively influences the model accuracy (what we show in Section~\ref{sec:exp}) and the model explanations, as some of the important prototypes can be removed only because of its similarity to the other class. That is why we provide the merge-pruning that can join quite a lot of semantically similar prototypes from different classes without significant change in accuracy. As an example, in Figure~\ref{fig:pruned_sup2}, we present eighteen prototypes from seventeen classes merged into one prototype corresponding to \textit{a belly brighter than the wings}.

\paragraph{Inter-class similarity discovery.}
The positive property of ProtoPShare is its ability to discover the similarity between classes based on the prototypes they share. Such similarity can be represented, i.a., by the graph that we present in Figure~\ref{fig:graph_cars} where each node corresponds to a class and the strength of edge between two nodes corresponds to the number of shared prototypes. Such analysis has many applications, like finding similar prototypical parts between two classes or clustering them into groups. Moreover, such a graph can be the starting point for more advanced visualizations that present prototypes shared between two classes after choosing one of the graph's edges.

\begin{figure}[t]
    \centering
    \includegraphics[width=0.5\textwidth]{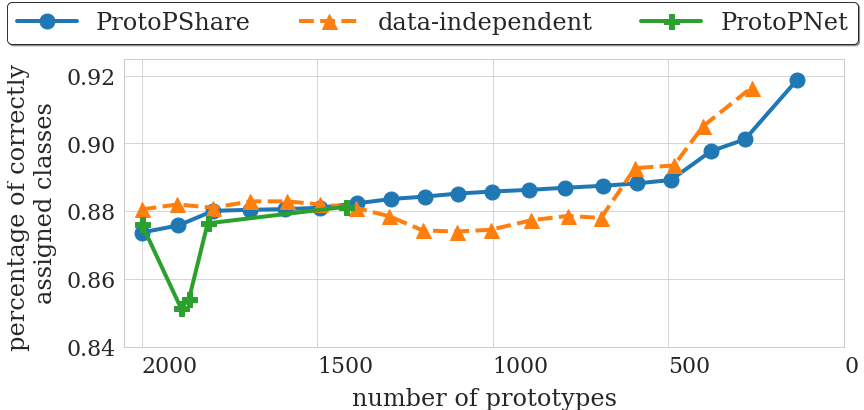}
    \caption{Percent of patches correctly assigned to their classes in the successive steps of the pruning (higher is better). The ProtoPShare pruning method is the only one that constantly increases this number, which means that it merges similar prototypes from the model perspective and is more effective in using the model capacity.}
    \label{fig:patches_per}
\end{figure}

\paragraph{Why data-dependent similarity?} To explain the superiority of our data-dependent similarity over the Euclidean norm, in Figure~\ref{fig:dd_di_similarity}, we present the pairs of prototypes close according to $d_{DD}$ similarity and distant in representation space. It can be noticed that the prototypes of one pair are semantically similar, even though they differ in colors' distribution. In our opinion, the ability to find such pairs of similar prototypes is the main advantage of our data-dependent similarity that distinguishes it from the Euclidean norm.
This ability is extremely important after the merge-pruning step because, as presented in Figure~\ref{fig:patches_per}, ProtoPShare representations of the image's parts more often activate the prototypes assigned to their true class, which indirectly results in higher accuracy and robustness. Notice that, to generate Figure~\ref{fig:patches_per}, we create a dataset with $5$ parts from each testing image with the highest prototype activation, and then check which of those activations corresponds to a true class. We conclude that only ProtoPShare constantly increases this number, which means that it merges similar prototypes from the model perspective and is more effective in using the model capacity.

\paragraph{User study.}
To further verify the superiority of our data-dependent pruning, we conducted user study where we presented two pairs of prototypes to the users. One of the pair contains prototypes most similar according to our data-dependent similarity, while the other contains prototypes closest in the representation space. The pairs were presented at once but in random order to prevent the order bias, and the users were asked to indicate a more consistent pair of prototypes (with an additional option ``Impossible to decide''). The study was conducted on $57$ users asked $5$ times for a different randomly chosen set of prototypes, resulting in $285$ answers altogether. Among all the answers, $195$ favored our approach, $61$ preferred data-independent similarity, and finally, in $29$ cases, users could not decide which similarity is better. Therefore, we conclude that our data-dependent similarity is considered as more consistent than its data-independent counterpart.

\paragraph{Discussion.}
In this section, we explained why prototype merge is preferred over background pruning and why it is crucial to use data-dependent similarity instead of the Euclidean norm (also with user study). Moreover, we showed how ProtoPShare can be used for similarity discovery.

\section{Conclusions}
\label{sec:dc}

We presented ProtoPShare, a self-explained method that incorporates the paradigm of prototypical parts to explain its predictions. The method extends the existing approaches because it can share the prototypes between classes, reducing their number up to three times. To efficiently share the prototypes, we introduced our data-dependent pruning that merges prototypes with similar semantics. As a result, we increased the model's interpretability and enabled similarity discovery while maintaining high accuracy, as we showed through theoretical results and many experiments, including user study.

{\small
\bibliographystyle{ieee_fullname}
\bibliography{egbib}
}

\maketitle
\onecolumn

This document represents supplementary materials to the paper ``ProtoPShare: Prototype Sharing for Interpretable Image Classification and Similarity Discovery''. It contains six sections. Section~\ref{sec:theorem} contains a proof to Theorem~1, Section~\ref{sec:training} describes training details, and Section~\ref{sec:pruning} illustrates the undesirable behavior of ProtoPNet pruning. Section~\ref{sec:similarity_distributions} presents the distribution of normalized distances between pairs of prototypes from ProtoPShare and its data-independent alternative, while Section~\ref{sec:user_study} in details describes the user study questionnaire. Finally, Section~\ref{sec:extended_fig} contains an extension of the figures and tables from the paper.

\section{Proof for Theorem 1}\label{sec:theorem}

In this section, we recall Theorem 1 and provide its proof omitted in the paper due to the page limit. It examines changes in the network's predictions after the prototypes' merge. 

\begin{thm}
Let $x\in X$ be an input image correctly classified by ProtoPShare $h\circ g_p\circ f$ as class $c$ before the prototypes' merge, and let: \vspace{-5pt}
\begin{enumerate}[label=(A\arabic*)]

    \item some of the class $k$ prototypes remain unchanged $P'_k\subset P_k$ while the others merge into the other classes' prototypes $S_k\subset\bigcup\limits_{i\neq k}P'_i$,
    
    \item $p\in P_k\setminus P'_k$ is merged into $\tilde{p}\in S_k$,

    \item $z_p=\argmin\limits_{z\in \bigcup\limits_{x\in X}Z_x} \|z - p\|$ is the representation of the training patch nearest to any prototype $p$, \label{assumption_nearest}

	\item there exist $\delta\in(0, 1)$ such that:
	\begin{enumerate}[label=\alph*), ref=(A\arabic{enumi}\alph*)]

    	\item for $k\neq c$, $p\in P_k\setminus P_k'$ and $\tilde{p}\in S_k$, we suppose that $\|p - \tilde{p}\| \leq\theta\|z_p - p\| - \sqrt{\varepsilon}$ and $\theta = \min(\sqrt{1 + \delta} - 1, 1 - \frac{1}{\sqrt{2-\delta}})$ ($\varepsilon$ comes from function $g$ defined in Section~3 of the paper, \label{assumption_ineq_a}

    	\item for $p\in P_c\setminus P'_c$ and $\tilde{p}\in S_c$, we suppose that $\|\tilde{p} - p\| \leq(\sqrt{1+\delta}-1)\|z_p - p\|$ and $\|z_p - p\|\leq\sqrt{1-\delta}$,
	\label{assumption_ineq_b}
	\end{enumerate}
    \vspace{-5pt}
	\item for each class $k$, weights connecting class with assigned prototypes equal $1$, and the other weights equal 0 (i.e., $w_h(p) = 1$ for $p\in P_k\cup S_k$ and $w_h(p) = 0$ for $p\in\bigcup\limits_{i\neq k}P_i\setminus S_k$). \label{assumption_weight}
\end{enumerate}
\vspace{-5pt}
\noindent Then after the prototypes' merge, the output logit for class $c$ can decrease at most by $\Delta^c_{\max}$, and the output logit for the other classes $k\neq c$ can increase at most by $\Delta^k_{\max}$, where:
\[
\Delta^k_{\max}:=|P_k\setminus P'_k|\log \left((1+\delta)(2-\delta)\right) \; \text{ for }\; k\in\{1, \ldots, K\}. \]
If the output logits between the top-2 classes are at least $\Delta^c_{\max} + \max\limits_{k\neq c}\big(\Delta^k_{\max}\big)$ apart, then the merge of prototypes does not change the prediction of $x$.
\end{thm}\bigskip

\begin{proof}
For any class $k$, let
\[
	L_k(x, P) = \sum_{p\in P} w_h(p)\cdot\log\left(\dfrac{\|z_p - p\|^2 + 1}{\|z_p - p\|^2 + \varepsilon}\right)
\]
denotes its output logit for an input image $x\in X$.
From assumption~\ref{assumption_weight}, before the merge, we have
\[
L_k(x, P_k) = \sum_{p\in P_k}\log\left(\dfrac{\|z_p - p\|^2 + 1}{\|z_p - p\|^2 + \varepsilon}\right),
\]
and after merge
\[
L_k(x, P'_k\cup S_k) = \sum_{p\in P'_k\cup S_k}\log\left(\dfrac{\|z_p - p\|^2 + 1}{\|z_p - p\|^2 + \varepsilon}\right).
\]

\noindent If none of the prototypes from class $k$ is merged into the other prototype (i.e., $S_k=\emptyset$), then there are no changes at the output (i.e., $L_k(x, P'_k\cup S_k) - L_k(x, P_k)=0$). In the opposite case
\begin{align*}
	L_k(x, P'_k\cup S_k) - L_k(x, P_k) & = L_k(x, S_k) - L_k(x, P_k\setminus P'_k)\\
	& = \sum_{p\in S_k}\log\left(\dfrac{\|z_p - p\|^2 + 1}{\|z_p - p\|^2 + \varepsilon}\right) - \sum_{p\in P_k\setminus P'_k}\log\left(\dfrac{\|z_p - p\|^2 + 1}{\|z_p - p\|^2 + \varepsilon}\right) \\
	& = \sum^{|P_k\setminus P'_k|}_{i=1} \log\left(\dfrac{\|z_{\tilde{p_i}} - \tilde{p_i}\|^2 + 1}{\|z_{p_i} - p_i\|^2 + 1}\cdot\dfrac{\|z_{p_i} - p_i\|^2 + \varepsilon}{\|z_{\tilde{p_i}} - \tilde{p_i}\|^2 + \varepsilon}\right).
\end{align*}
For each class $k\in\{1,.., K\}$ and its prototypes $i\in\{1,..,|P_k\setminus P'_k|\}$, let
\[
	\vartheta_i := \dfrac{\|z_{\tilde{p_i}} - \tilde{p}_i\|^2 + 1}{\|z_{p_i} - p_i\|^2 + 1}\cdot\dfrac{\|z_{p_i} - p_i\|^2 + \varepsilon}{\|z_{\tilde{p_i}} - \tilde{p}_i\|^2 + \varepsilon}.
\]\bigskip

\noindent To obtain the lower bound of $\vartheta_i$ for $k=c$ (correct class of $x$), we use the second inequality in~\ref{assumption_ineq_b}, receiving
\[
	\frac{\|z_{\tilde{p_i}} - \tilde{p}_i\|^2 + 1}{\|z_{p_i} - p_i\|^2 + 1} \geq \frac{1}{\|z_{p_i} - p_i\|^2 + 1} \overset{by~\ref{assumption_ineq_b}}{\geq} \frac{1}{2 - \delta}.
\]
By the assumption~\ref{assumption_nearest} and the triangle inequality we obtain
\(
    \|z_{\tilde{p_i}} - \tilde{p}_i\|\leq \|z_{p_i} - \tilde{p}_i\|\leq \|z_{p_i} - p_i\| + \|\tilde{p}_i - p_i\|.
\)
From the assumption~\ref{assumption_ineq_b} we obtain
\begin{equation}\label{ineq_a}
 \|\tilde{p}_i - p_i\|\overset{by~\ref{assumption_ineq_b}}{\leq} (\sqrt{1+\delta} - 1)\|z_{p_i} - p_i\| \implies \|\tilde{p}_i - p_i\| + \|z_{p_i} - p_i\|\leq \sqrt{1+\delta}\|z_{p_i} - p_i\|. 
\end{equation}
Hence
\[
	\frac{\|z_{p_i} - p_i\|^2 + \varepsilon}{\|z_{\tilde{p}_i} - \tilde{p}_i\|^2 + \varepsilon} \geq \frac{\|z_{p_i} - p_i\|^2 + \varepsilon}{(\|z_{p_i} - p_i\| + \|\tilde{p}_i - p_i\|)^2 + \varepsilon} \overset{by~\eqref{ineq_a}}{\geq} \frac{1}{1 + \delta}.
\]
Combining the above inequalities, we have
\[
	\vartheta_i = \frac{\|z_{\tilde{p}_i} - \tilde{p}_i\|^2 + 1}{\|z_{p_i} - p_i\|^2 + 1}\cdot\frac{\|z_{p_i} - p_i\|^2 + \varepsilon}{\|z_{\tilde{p}_i} - \tilde{p}_i\|^2 + \varepsilon} \geq \frac{1}{(1+\delta)(2 - \delta)}.
\]
It means that the output logit change of class $c$ as a result of the prototype merge satisfies
\[
    L_c(x, P'_c\cup S_c) - L_c(x, P_c) = \sum^{|P_c\setminus P'_c|}_{i=1} \log\vartheta_i \geq -|P_c\setminus P'_c|\log((1+\delta)(2-\delta))
\]
or equivalently, $L_c(x, P_c) - L_c(x, P'_c\cup S_c)\leq|P_c\setminus P'_c|\log((1+\delta)(2-\delta))$. Hence, the worst decrease of the class $c$ output logit as a result of prototype merge is $\Delta^c_{\max}$.\bigskip

\noindent To obtain the lower bound of $\vartheta_i$ for $k\neq c$ (incorrect class of $x$), we use the triangle inequality together with  assumptions~\ref{assumption_nearest} and~\ref{assumption_ineq_a}, receiving
\begin{align*}
    \|z_{\tilde{p}_i} - \tilde{p}_i\|^2 & \overset{by~\ref{assumption_nearest}}{\leq} \|z_{p_i} - \tilde{p}_i\|^2 \leq(\|z_{p_i} - p_i\| + \|\tilde{p}_i - p_i\|)^2 \\
    & \overset{by~\ref{assumption_ineq_a}}{\leq} (\|z_{p_i} - p_i\| + (\sqrt{1+\delta} - 1)\|z_{p_i} - p_i\|)^2 = (1 + \delta)\|z_{p_i} - p_i\|^2.
\end{align*}
Hence
\begin{equation}\label{ineq_1}
    \frac{\|z_{\tilde{p}_i} - \tilde{p}_i\|^2 + 1}{\|z_{p_i} - p_i\|^2 + 1}  \leq\frac{(1 + \delta)\|z_{p_i} - p_i\|^2 + 1}{\|z_{p_i} - p_i\|^2 + 1}\leq 1 + \delta.
\end{equation}\bigskip

\noindent To derive an upper bound for the second part of $\vartheta_i$, we use the assumption~\ref{assumption_nearest} and~\ref{assumption_ineq_a}
\[
\|p_i - \tilde{p}_i\| \overset{by~\ref{assumption_ineq_a}}{\leq} \theta\|z_{p_i} - p_i\| -\sqrt{\varepsilon}
\overset{by~\ref{assumption_nearest}}{\leq} \theta\|z_{\tilde{p}_i} - p_i\| - \sqrt{\varepsilon} \overset{by~\ref{assumption_ineq_a}}{\leq} \left(1 - \frac{1}{\sqrt{2 - \delta}}\right)\|z_{\tilde{p}_i} - p_i\| - \sqrt{\varepsilon}.
\]
From which we know that $\|z_{\tilde{p}_i} - p_i\| - \|\tilde{p}_i - p_i\| > 0$. Again, we use the triangle inequality
\[
\|z_{\tilde{p}_i} - p_i\| \leq \|z_{\tilde{p}_i} -\tilde{p}_i\| + \|\tilde{p}_i - p_i\| \implies \|z_{\tilde{p}_i} - \tilde{p}\|\geq \|z_{\tilde{p}_i} - p_i\| - \|\tilde{p}_i - p_i\|.
\] 
By reusing assumptions~\ref{assumption_nearest} and~\ref{assumption_ineq_a}, we have
\begin{equation}\label{ineq_star}
    \frac{\|z_{p_i} - p_i\| + \sqrt{\varepsilon}}{\sqrt{2 -\delta}} \leq \frac{1}{\sqrt{2 - \delta}}\|z_{p_i} - p_i\| + \sqrt{\varepsilon} \overset{\text{by~\ref{assumption_ineq_a}}}{\leq} \|z_{p_i} - p_i\| - \|\tilde{p}_i - p_i\| \overset{\text{by~\ref{assumption_nearest}}}{\leq} \|z_{\tilde{p}_i} - p_i\| - \|\tilde{p}_i - p_i\|.
\end{equation}
Thanks to the above, we get
\begin{equation}\label{ineq_2}
    \frac{\|z_{p_i} - p_i\|^2 + \varepsilon}{\|z_{\tilde{p}_i} - \tilde{p}_i\|^2 + \varepsilon} \leq \frac{\|z_{p_i} - p_i\|^2 + \varepsilon}{(\|z_{\tilde{p}_i} - p_i\| - \|\tilde{p}_i - p_i\|)^2 + \varepsilon} 
    \leq \left(\frac{\|z_{p_i} - p_i\| + \varepsilon}{\|z_{\tilde{p}_i} - p_i\| - \|\tilde{p}_i - p_i\|}\right)^2 \overset{by~\eqref{ineq_star}}{\leq} 2 - \delta.
\end{equation}
It means that the output logit change of class $k$ as a result of the prototype merge satisfies
\[
L_k(x, P'_k\cup S_k) - L_k(x, P_k) = \sum^{|P_k\setminus P'_k|}_{i=1} \log\vartheta_i \overset{\text{by~\eqref{ineq_1}, \eqref{ineq_2}}}{\leq} |P_k\setminus P'_k|\log((1+\delta)(2-\delta)).
\]
Hence, the worst increase of the class $k$ output logit as a result of prototype merge is $\Delta^k_{\max}$.\bigskip

\noindent To prove the last thesis, let us assume that the output logit $L_c(x, P_k)$ of the correct class $c$ before prototype merge is at least $\Delta^c_{\max} + \max\limits_{k\neq c}\big(\Delta^k_{\max}\big)$ higher than the output logit $L_k(x, P_k)$ of any other class $k\neq c$, i.e.,
\begin{equation} \label{assumption_thesis2}
 L_c(x, P_c) \geq L_k(x, P_k) + \Delta^c_{\max} + \max\limits_{k\neq c}\big(\Delta^k_{\max}\big) \quad\text{for}\quad k\neq c.
\end{equation}
Since the output logit of the correct class $c$ satisfies
\begin{equation} \label{output_correct_class}
 L_c(x, P'_c\cup S_c) \geq L_c(x, P_c) - \Delta^c_{\max}
\end{equation}
and the output logit of any class $k\neq c$ satisfies
\begin{equation} \label{output_incorrect_class}
 L_k(x, P'_k\cup S_k) \leq L_k(x, P_k) + \Delta^k_{\max},
\end{equation}
for any $k\neq c$ we have
\begin{align*}
    L_c(x, P'_c\cup S_c) & \overset{by~\eqref{output_correct_class}}{\geq} L_c(x, P_c) - \Delta^c_{\max}
    \overset{by~\eqref{assumption_thesis2}}{\geq} L_k(x, P_k) + \max\limits_{k\neq c}\big(\Delta^k_{\max}\big)
    \overset{by~\eqref{output_incorrect_class}}{\geq} L_k(x, P'_k\cup S_k).
\end{align*}
Hence, the input image $x$ will still be correctly classified as class $c$ after prototype merge.
\end{proof}

\section{Training details}\label{sec:training}
\setcounter{figure}{0}  

We decided to omit the training details in the paper for its clarity. However, to make the paper reproducible, we provide them in this section.

First of all, we perform exhaustive offline data augmentation using rotation, skewing, flipping, and shearing with the probability of $0.5$ for each operation. When it comes to the architecture, $f$ is followed by two additional $1\times 1$ convolutional layers before the prototypes' layer, and we use ReLU as the activation function for all convolutional layers except the last one (with the sigmoid function). The training bases on the images cropped (using the bounding-boxes provided with the datasets) and resized to $224 \times 224$ pixels. As a result, we obtain a convolutional feature map of size $7 \times 7 \times 256$. Hence, prototypes have size $1 \times 1 \times 256$.
During training, we use Adam optimizer~\cite{kingma2014adam} with $\beta_1=0.9$ and $\beta_2=0.999$, batch size $80$, learning rates $10^{-4}$ for $f$ and $h$, and $3\cdot 10^{-3}$ for additional $1\times 1$ convolutions and prototypes' layer. For the loss function defined in~\cite{chen2019looks}, we use weights $\lambda_1=0.8$, $\lambda_2=0.08$, and $\lambda_l=10^{-4}$. Moreover, the learning rates for convolutional parts and prototype layer are multiplied by $0.1$ every $5$ epochs of the training.

\section{Undesirable behaviors of ProtoPNet pruning}\label{sec:pruning}
\setcounter{figure}{0}  

\begin{figure}[h]
    \centering
    \includegraphics[width=0.95\textwidth]{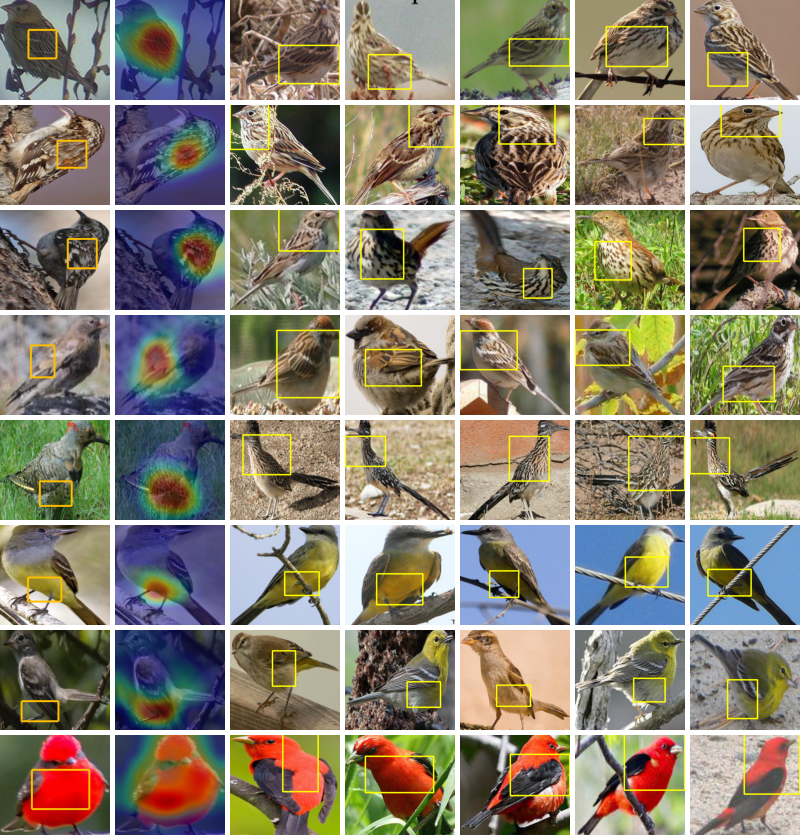}
    \caption{Example prototypes (one per row) pruned by ProtoPNet that do not represent the background. The first column corresponds to the image with the prototypical part (in a yellow bounding box). The second column shows the prototype's activation map for this image obtained as described in~\cite{chen2019looks}. The remaining five columns present images' parts (in yellow bounding boxes) whose representations are closest to the considered prototype. We observe that none of the prototypes represent the background what illustrates that ProtoPNet prunes many meaningful prototypes.}
    \label{fig:small_removed}
\end{figure}

\begin{figure}[h]
    \centering
    \includegraphics[width=0.95\textwidth]{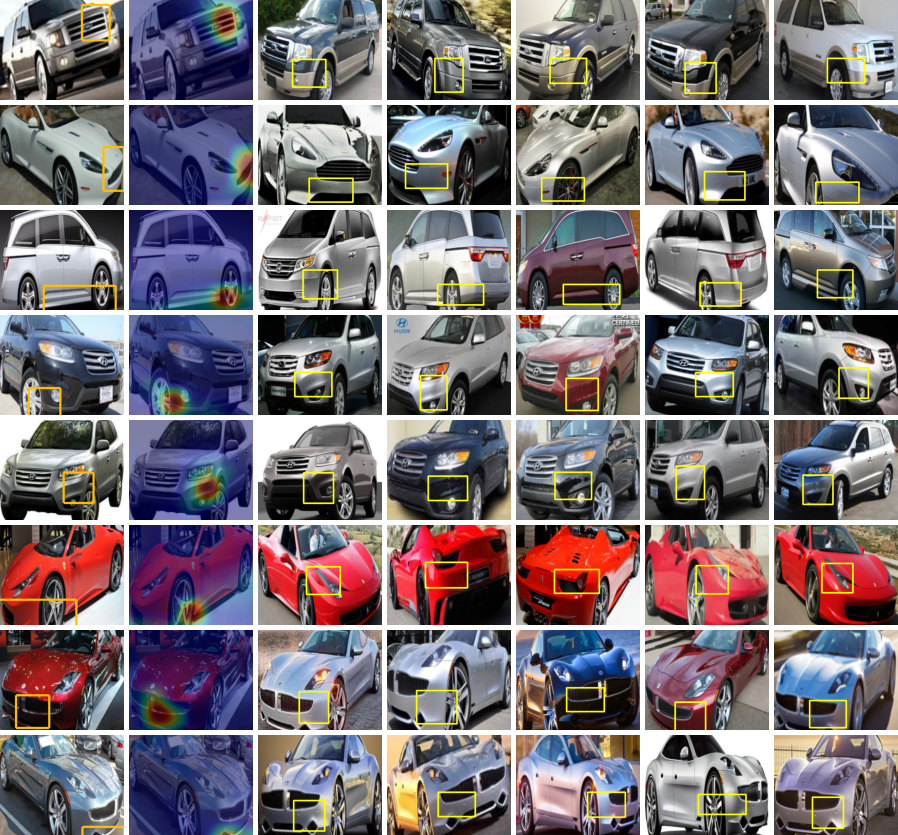}
    \caption{Example prototypes (one per row) pruned by ProtoPNet that do not represent the background. The first column corresponds to the image with the prototypical part (in a yellow bounding box). The second column shows the prototype's activation map for this image obtained as described in~\cite{chen2019looks}. The remaining five columns present images' parts (in yellow bounding boxes) whose representations are closest to the considered prototype. We observe that none of the prototypes represent the background what illustrates that ProtoPNet prunes many meaningful prototypes.}
    \label{fig:small_removed_cars}
\end{figure}

As we write in the paper, we decided to introduce our merge-pruning algorithm after a detailed analysis of the prototypes pruned by ProtoPNet and concluding that around $30\%$ of them represent significant prototypical parts instead of a background. This observation was consistent with Section 3.2 in~\cite{chen2019looks}, which states that ``the prototypes whose nearest training patches have mixed class identities usually correspond to background patches, and they can be automatically pruned from our model.'' Nevertheless, we were surprised by the high number of meaningful prototypes that are pruned. To illustrate the problem, we present some of them in Figure~\ref{fig:small_removed} and~\ref{fig:small_removed_cars} for the CUB-200-2011 and the Stanford Cars datasets, respectively.

\section{Similarity distributions}\label{sec:similarity_distributions}
\setcounter{figure}{0}

\begin{figure}[h]
    \centering
    \includegraphics[width=\textwidth]{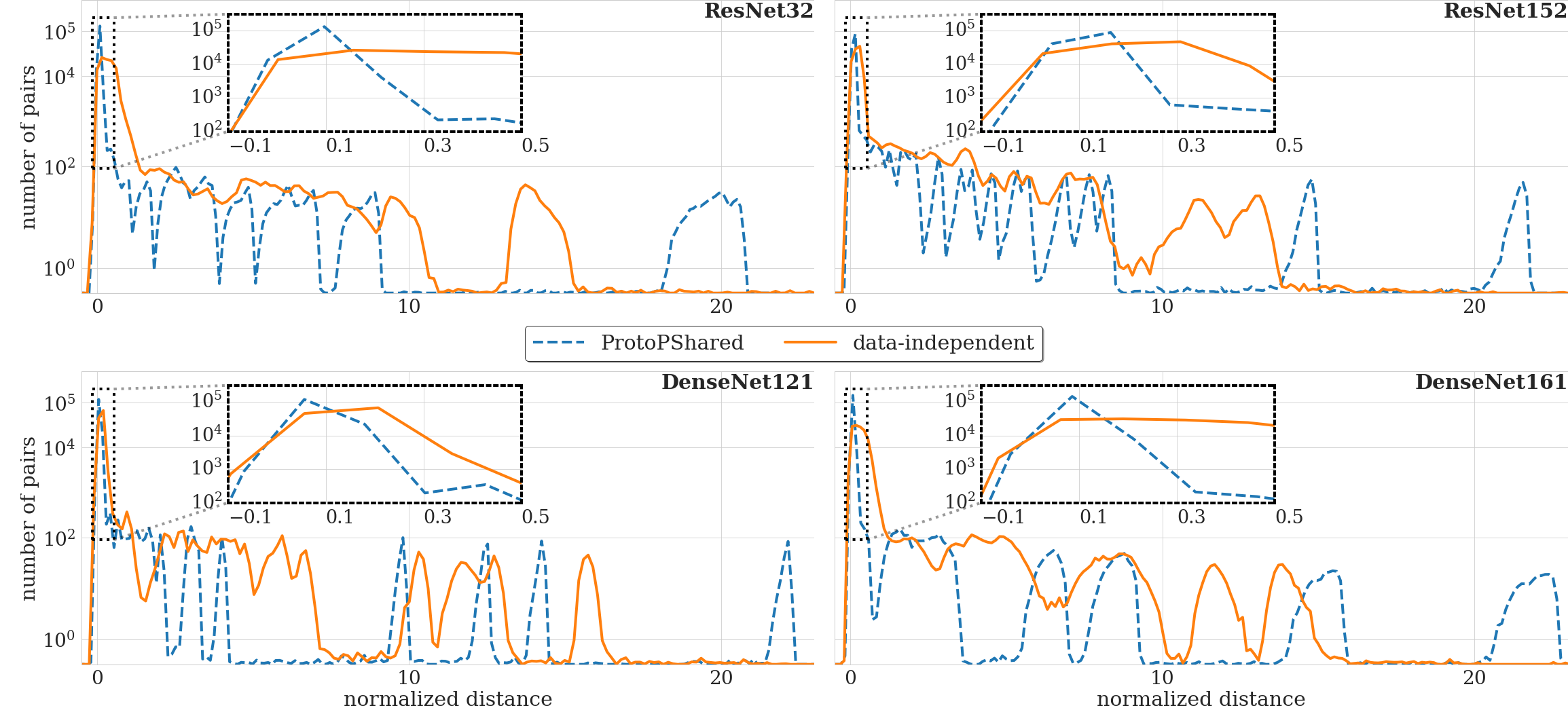}
    \caption{Distribution of normalized distances between pairs of prototypes from ProtoPShare and its data-independent alternative obtained for the CUB-200-2011 dataset. One can observe the peak at the beginning of data-dependent similarity distribution that results from finding similar concepts even if they are distant in representation space. Notice that the y-axis is in the logarithmic scale and that the initial parts of distributions are zoomed.}
    \label{fig:hist}
\end{figure}

One of the key novelty in our paper is introduction of the data-dependent similarity that finds semantically similar prototypes even if they are distant in the representation space. To further support this statement, in Figure~\ref{fig:hist}, we provide distributions of normalized distances between pairs of prototypes from ProtoPShare and its data-independent alternative obtained for the CUB-200-2011 dataset. The normalized distribution was prepared by computing the distances between all the prototypes' pairs, standardizing them (to have a mean of zero and a standard deviation of 1), and generating the histogram. One can observe the peak at the beginning of the data-dependent similarity distribution that does not appear in its data-independent counterpart. We hypothesize that this peak corresponds to the pairs of semantically similar prototypes distant from each other in the representation space due to various reasons, like differences in color distribution.

\section{User study questionnaire}\label{sec:user_study}
\setcounter{figure}{0}

\begin{figure}[h]
    \centering
    \includegraphics[width=\textwidth]{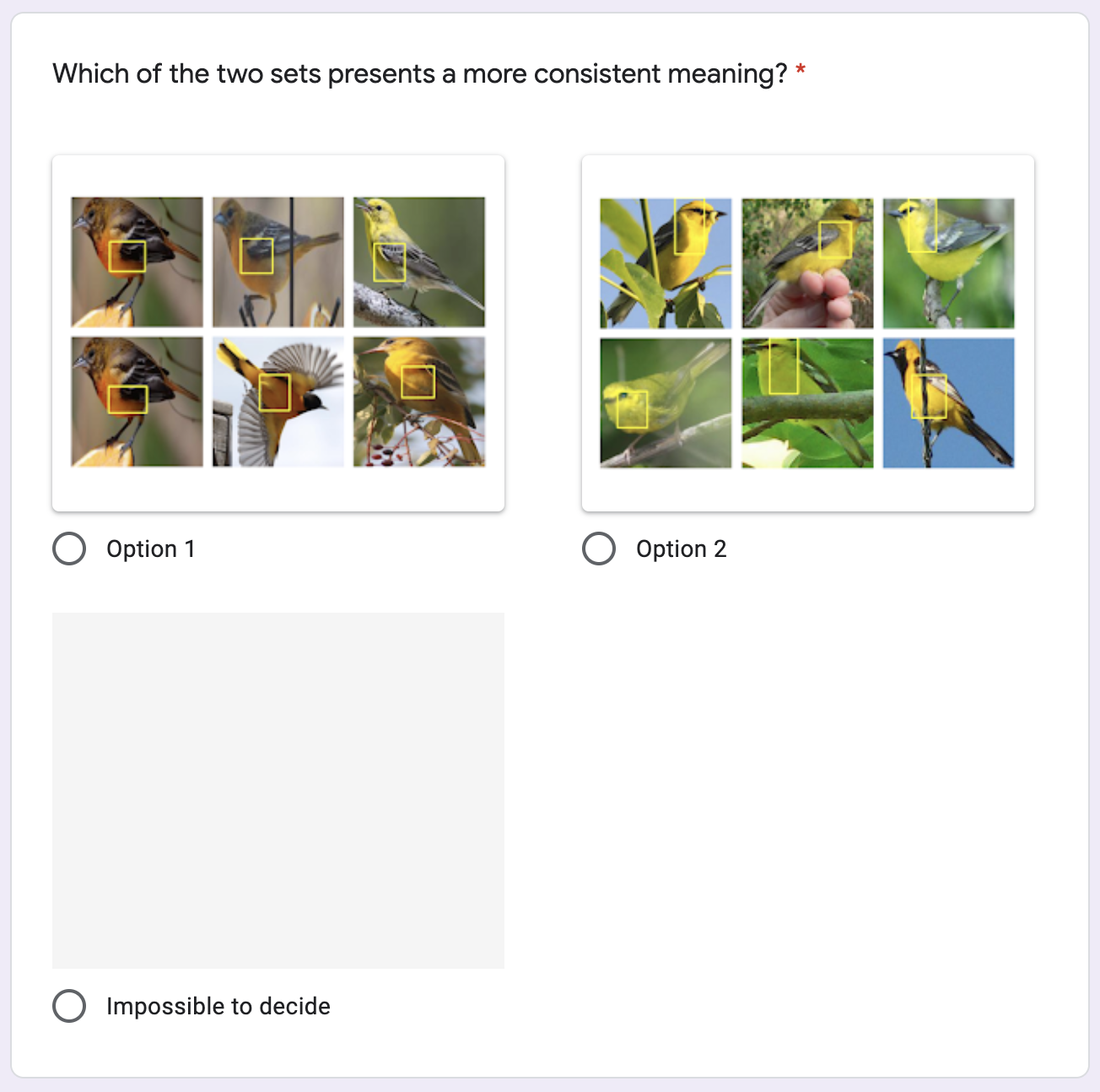}
    \caption{Example question from our user study.}
    \label{fig:survey}
\end{figure}

To further verify the superiority of our data-dependent pruning, we conduct a user study where we present five questions to the user. There were four versions of the questionnaire with different sets of questions. Each questionnaire had the following initial instruction: ``In this form, you will see two sets of images with yellow bounding-boxes. All bounding-boxes of one set should present one semantic meaning that corresponds to a particular characteristic of a bird's part (e.g., white head, dark neck, striped wing, or black legs). Please, decide which of the two sets presents more consistent bounding-boxes. This task can be difficult, as two sets you compare can have a different meaning. Nevertheless, try to decide which of them is more consistent, and if it is impossible, please choose the option Impossible to decide.'' After the initial instruction, we asked five times, ``Which of the two sets presents a more consistent meaning?''. An example question is presented in Figure~\ref{fig:survey}. We present the detailed results of the study in the paper.

\section{Extended versions of figures and tables}\label{sec:extended_fig}
\setcounter{figure}{0}

\begin{figure}[h]
    \centering
    \includegraphics[width=\textwidth]{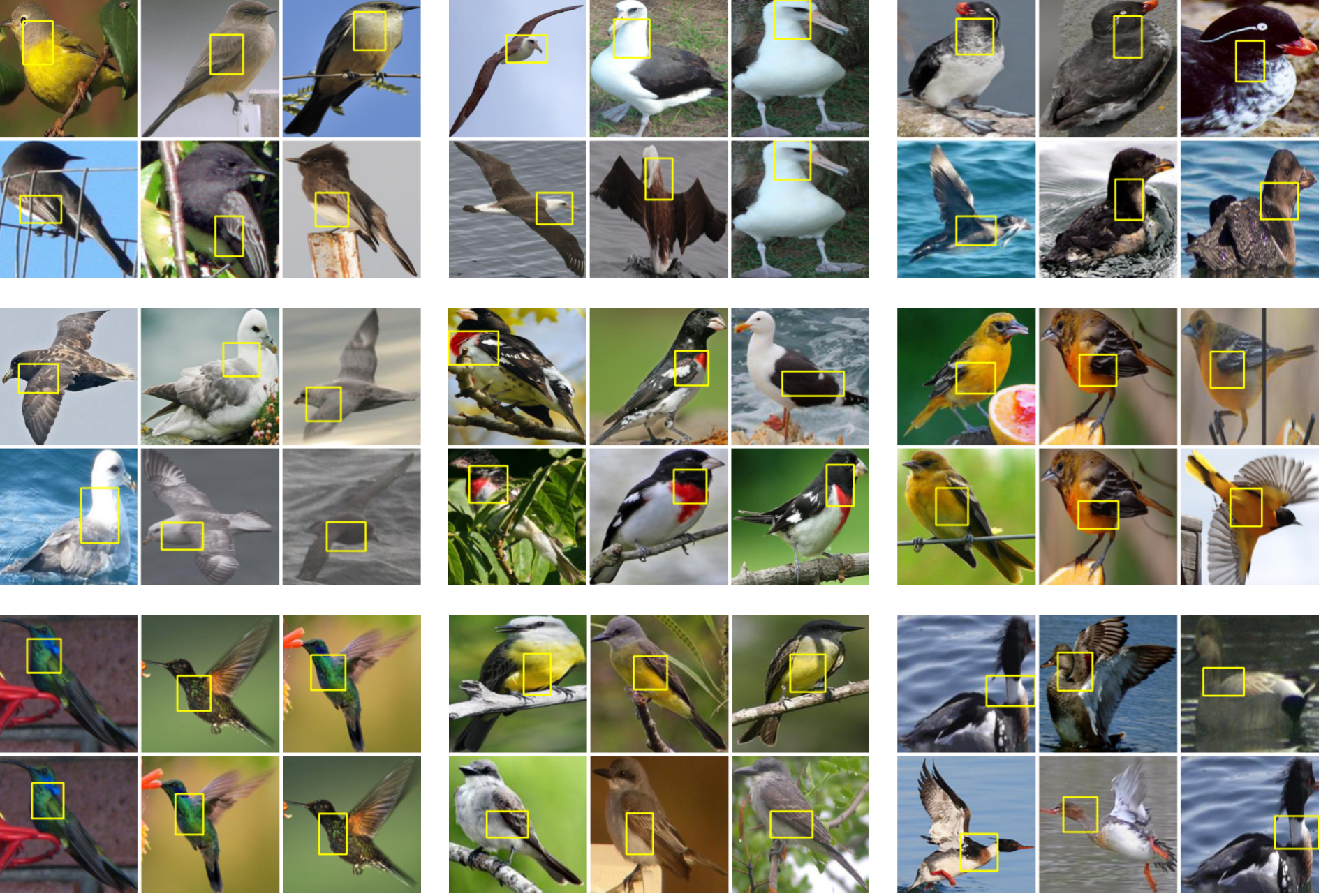}
    \caption{Pairs of semantically similar prototypes distant in the representation space but close according to our data-dependent similarity. Notice that each prototype is represented by one row of $3$ closest images’ parts (marked with yellow bounding-boxes), so each pair corresponds to $6$ images.}
    \label{fig:DD_similar}
\end{figure}

\begin{figure}[h]
    \centering
    \includegraphics[width=\textwidth]{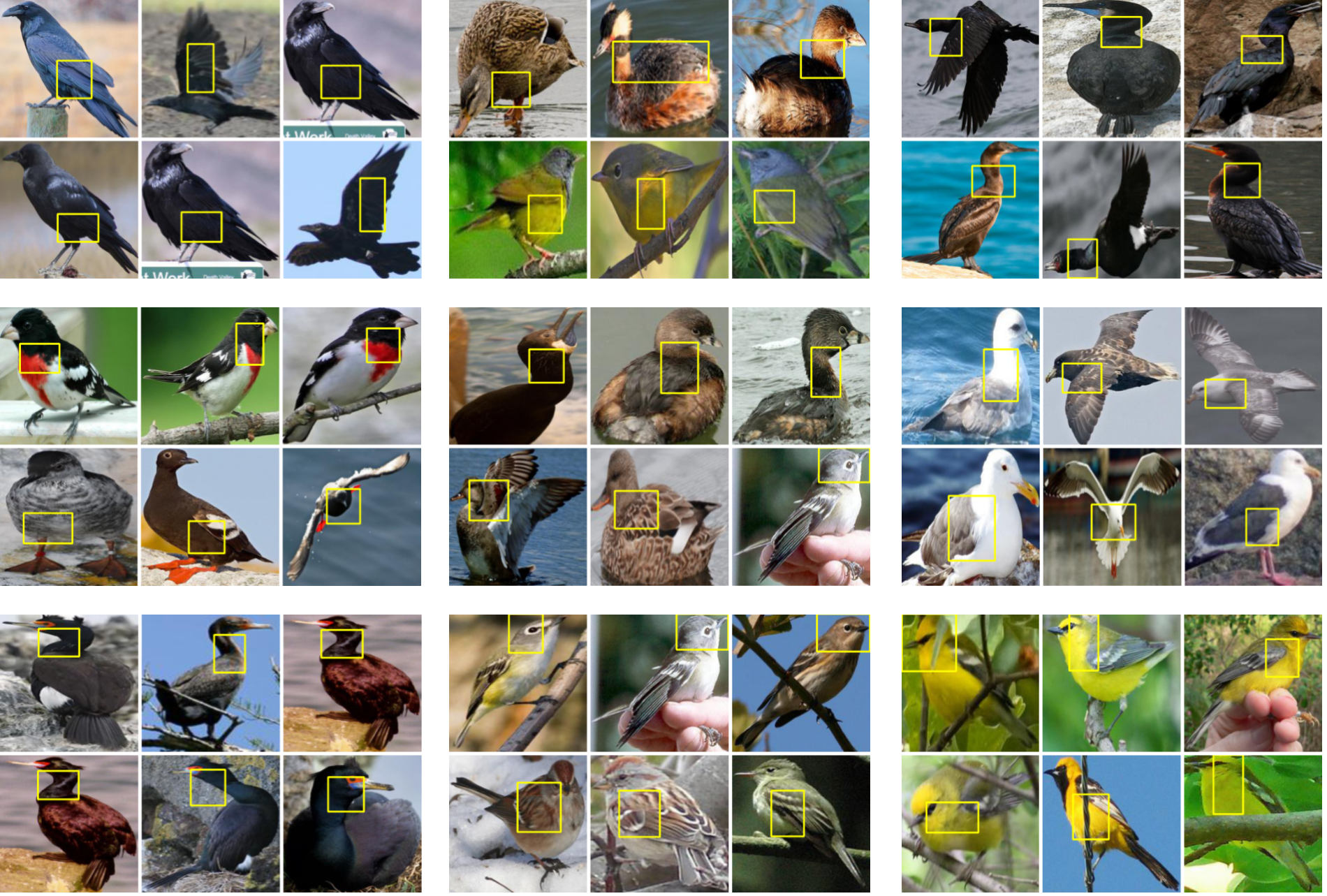}
    \caption{Pairs of prototypes close in the representation space. Notice that each prototype is represented by one row of $3$ closest images’ parts (marked with yellow bounding-boxes), so each pair corresponds to $6$ images.}
    \label{fig:DI_Similar}
\end{figure}

\begin{figure}[h]
    \centering
    \includegraphics[width=\textwidth]{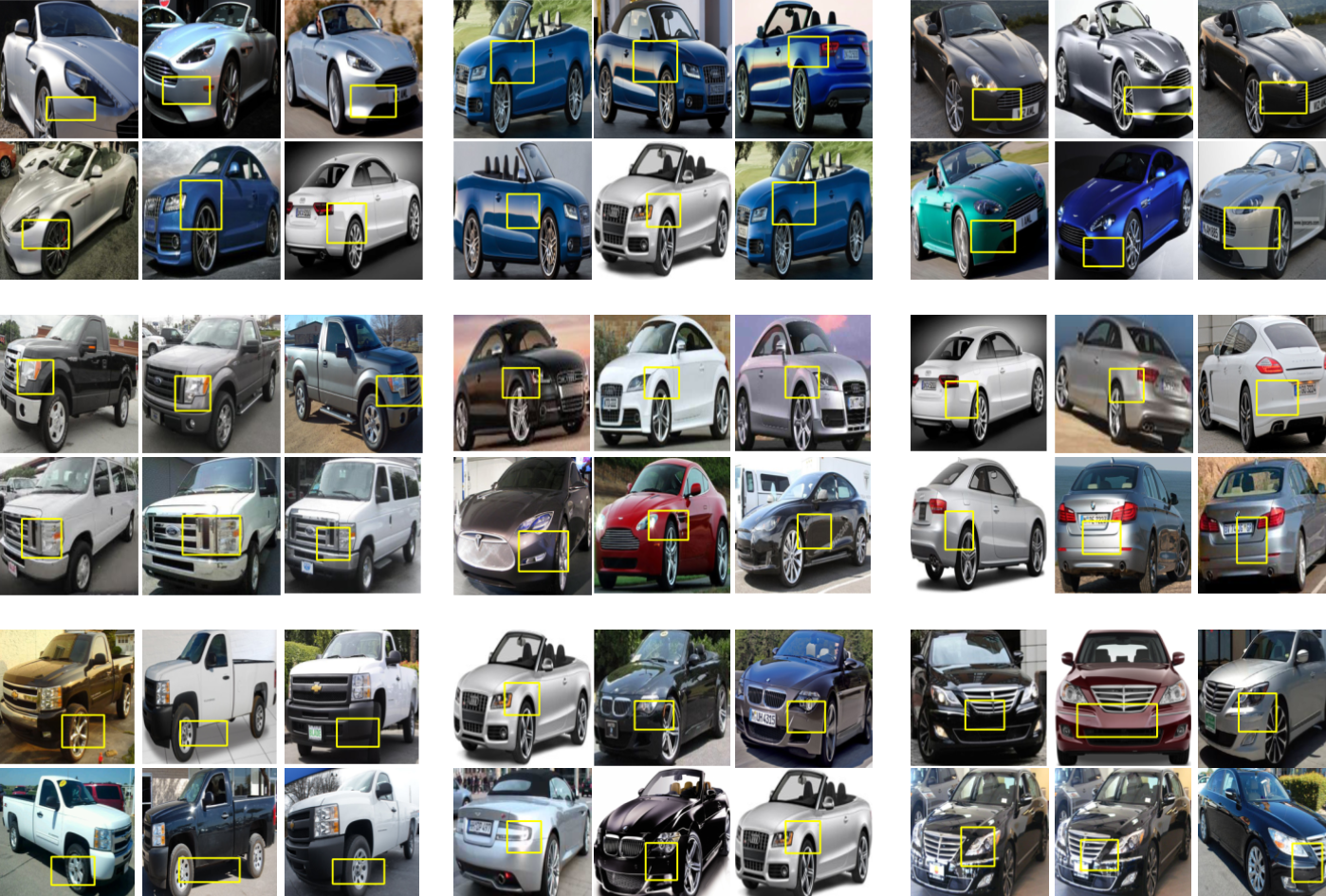}
    \caption{Pairs of semantically similar prototypes distant in the representation space but close according to our data-dependent similarity. Notice that each prototype is represented by one row of $3$ closest images’ parts (marked with yellow bounding-boxes), so each pair corresponds to $6$ images.}
    \label{fig:DD_Similar_cars}
\end{figure}

\begin{figure}[h]
    \centering
    \includegraphics[width=\textwidth]{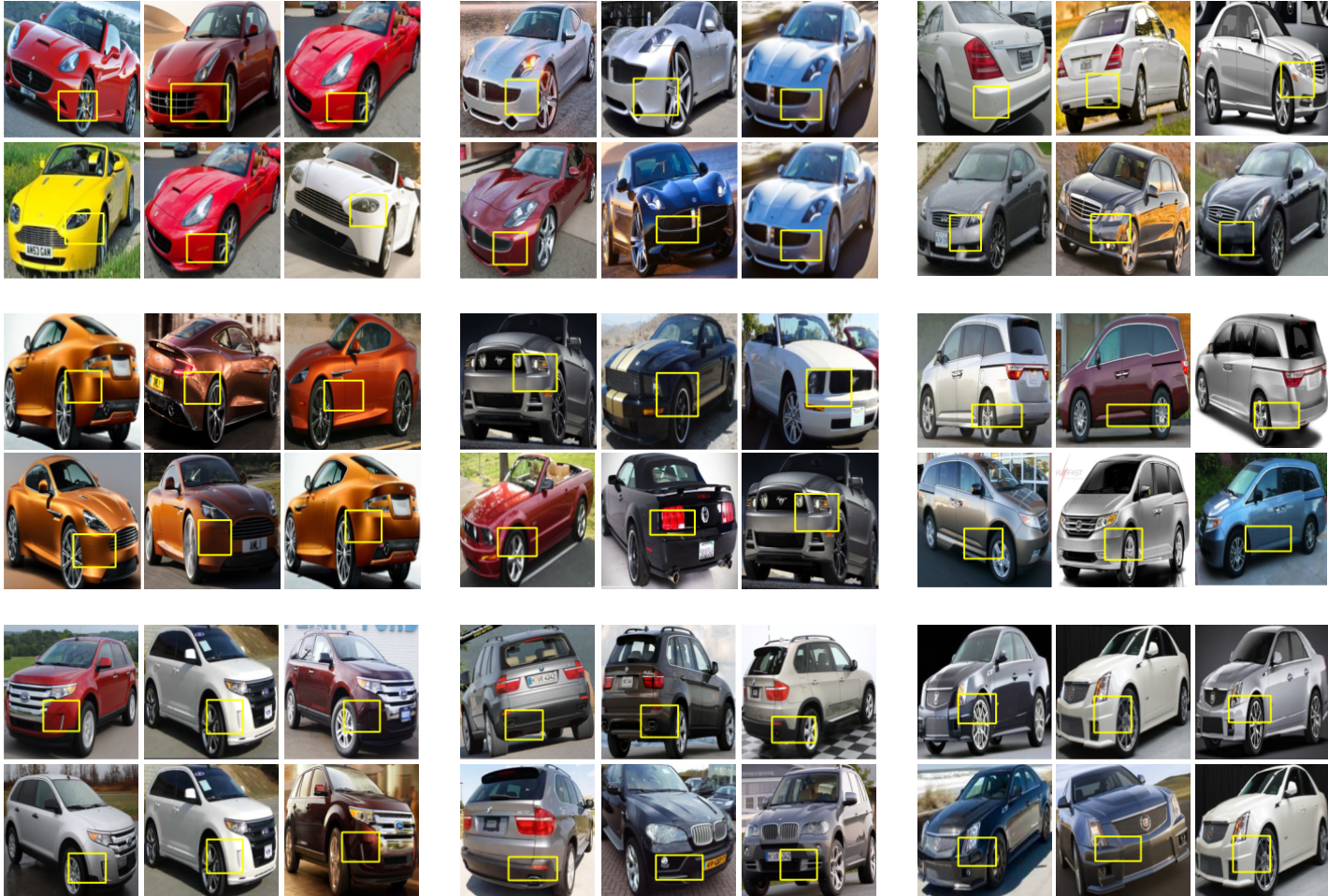}
    \caption{Pairs of prototypes close in the representation space. Notice that each prototype is represented by one row of $3$ closest images’ parts (marked with yellow bounding-boxes), so each pair corresponds to $6$ images.}
    \label{fig:DI_Similar_cars}
\end{figure}

\begin{figure}[h]
    \centering
    \includegraphics[width=0.47\textwidth]{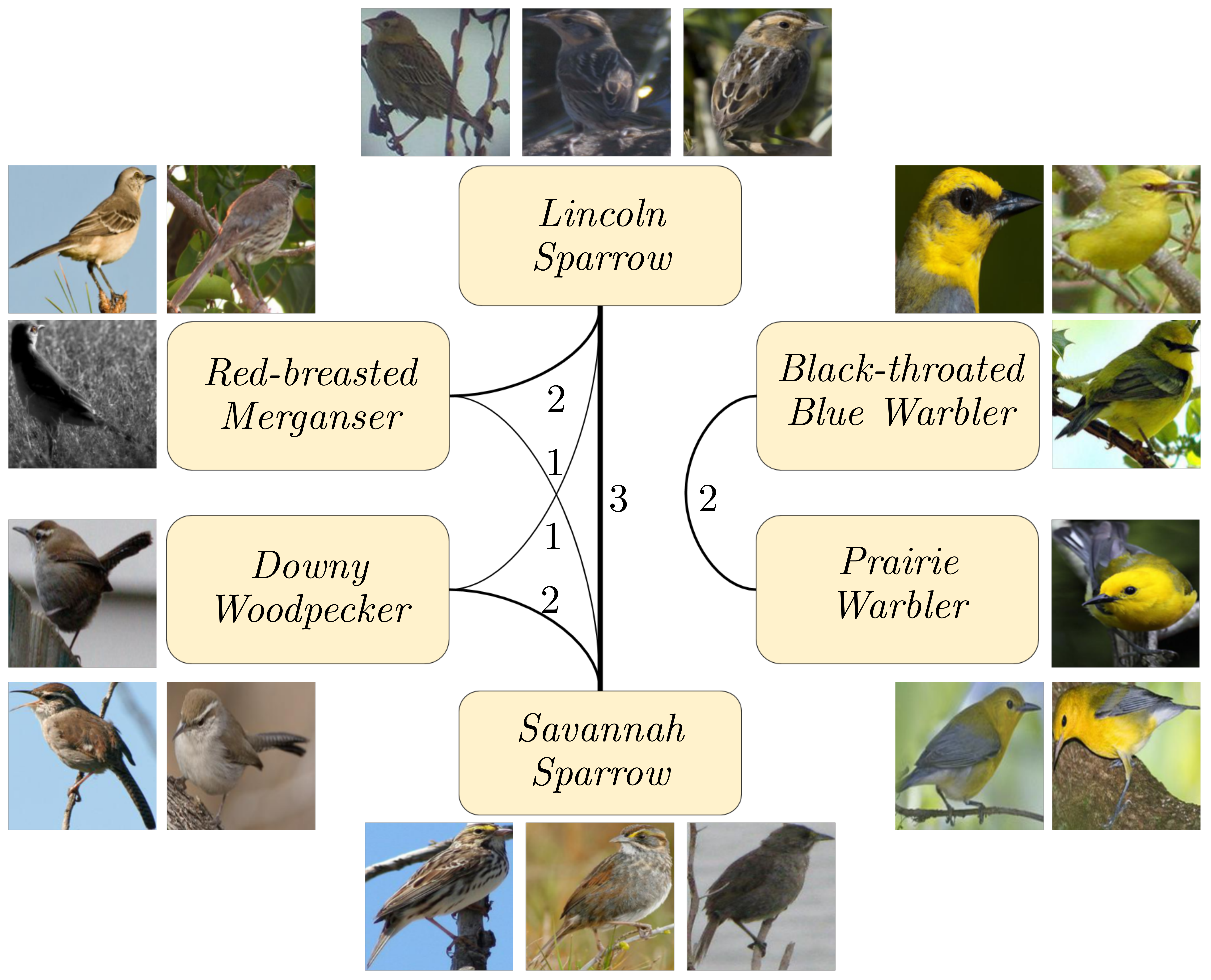}
    \caption{Inter class similarity visualization as a graph generated based on the prototypes shared in ProtoPShare where each node corresponds to a class and the strength of edge between two nodes corresponds to the number of shared prototypes. E.g., \textit{Lincoln sparrow} shares $3$ prototypes with \textit{Savannah sparrow} but does not share prototypes with \textit{Prairie warbler}. This graph can be used to find similarities between the classes or to cluster them into groups. Notice that each class is represented by three images located around the node.}
    \label{fig:graph}
\end{figure}

In this final section, we provide the extended version of the figures and tables from the paper.
In Figure~\ref{fig:DD_similar}, we present pairs of semantically similar prototypes distant in the representation space but close according to our data-dependent similarity. In Figure~\ref{fig:DI_Similar}, we present pairs of prototypes close in the representation space.
Figure~\ref{fig:DD_Similar_cars} and~\ref{fig:DI_Similar_cars} contains similar pairs but for the Stanford Cars dataset.
In Figure~\ref{fig:graph}, we provide inter-class similarity graph for CUB-200-2011 dataset.
In Figure~\ref{fig:prune_diff}, we compare methods accuracy for various pruning rates and all architectures on CUB-200-2011 dataset.
In Figure~\ref{fig:percents}, we compare methods accuracy for different percentages of prototypes merged per pruning step and all architectures on the same dataset.
Finally, in Table~\ref{tab:diff_optims}, we compare ProtoPShare to the other methods based on the prototypes’ for the Stanford Cars dataset.



\begin{figure}[h]
    \centering
    \includegraphics[width=\textwidth]{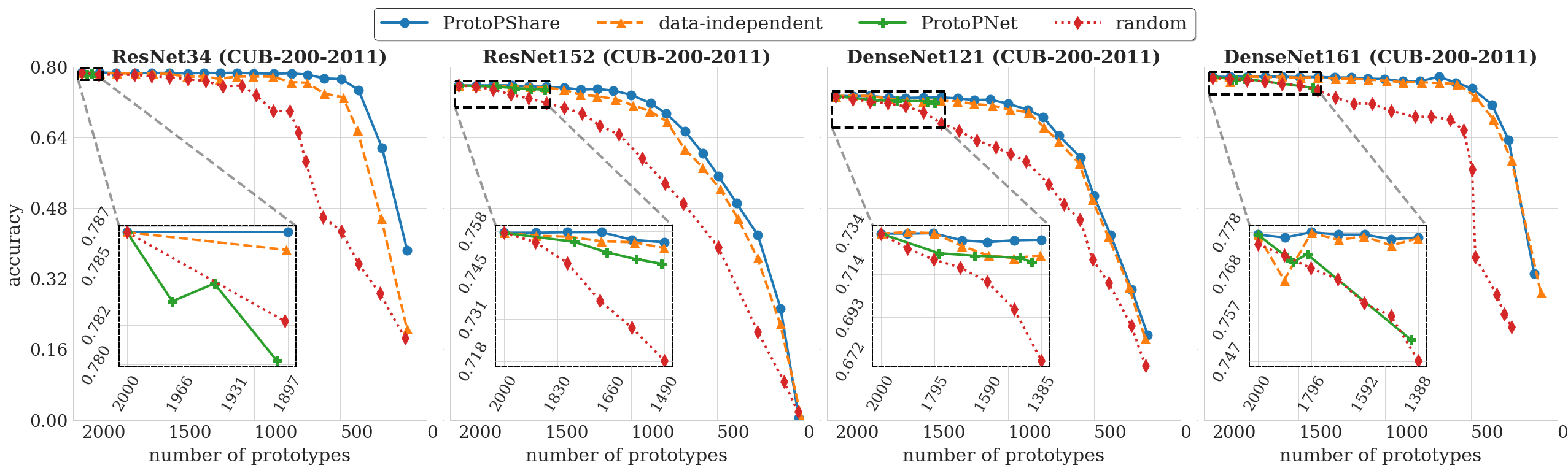}
    \caption{We compare the accuracy (higher is better) of ProtoPShare, ProtoPNet, and variations of our method (data-independent and random) for various pruning rates and all architectures on CUB-200-2011 dataset. Notice that the accuracies for the initial pruning steps are zoomed to enable comparison with ProtoPNet that can prune at most $30\%$ of prototypes.}
    \label{fig:prune_diff}
\end{figure}

\begin{figure}[h]
    \centering
    \includegraphics[width=\textwidth]{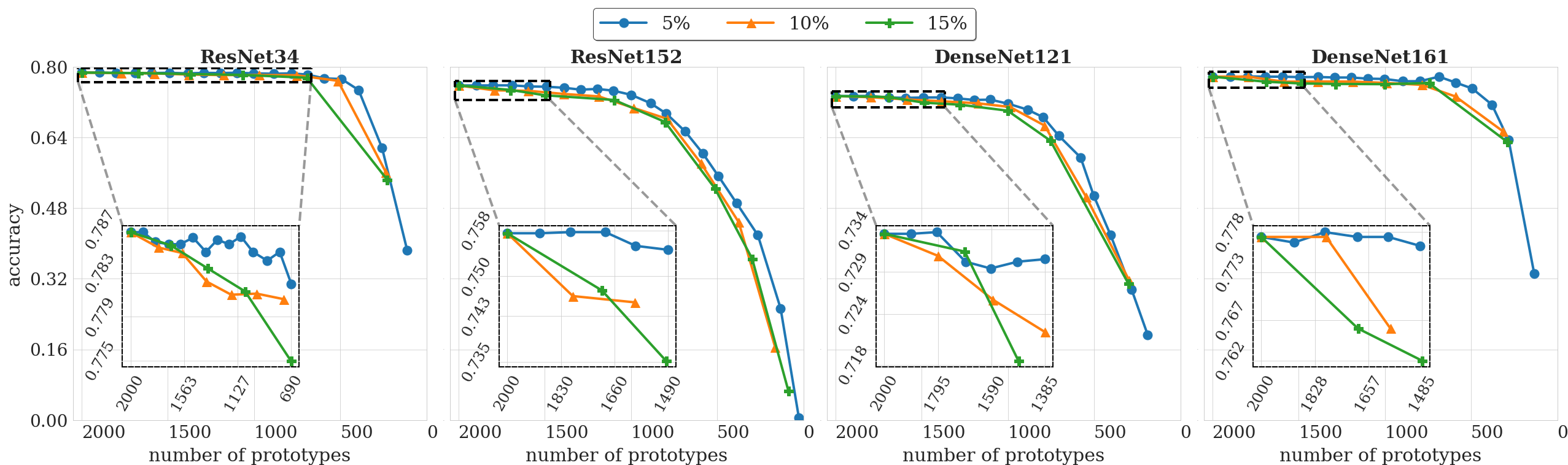}
    \caption{We compare the accuracy (higher is better) of ProtoPShare for different percentages of prototypes merged per pruning step ($5\%$, $10\%$, and $15\%$) for all architectures on the CUB-200-2011 dataset. Notice that the accuracy for the initial pruning steps is zoomed. The remaining architectures' results are presented in the Supplementary Materials.}
    \label{fig:percents}
\end{figure}

\begin{table}[htbp]\footnotesize
\setlength{\tabcolsep}{4pt}
\settowidth{\wexp}{Exponential}
\newcolumntype{C}{>{\centering\arraybackslash}p{\dimexpr.5\wexp-\tabcolsep}}

\hspace{0.25\textwidth}
    \begin{tabular*}{\columnwidth}{l@{\;\;}CCCCc@{\;\;}c@{\;\;}c@{\;\;}c@{}}
    \cmidrule[1pt](){1-9}
    \multirow{3}{*}{Model} & $|P|$ &  \multicolumn{3}{c}{Finetuning after} & \makecell{RN34} &  \makecell{DN121}  \\
     & before & \multicolumn{3}{c}{\makecell{pruning step}} & \multicolumn{4}{c}{$|P|$ in the final model} \\
     & pruning & $f$ & $P$ & $h$ & 480 & 980  \\
    \cmidrule(){1-9}
    ProtoPNet & & \multicolumn{3}{c}{no pruning} & 0.8474 & 0.7835 \\
    shared in training & & \multicolumn{3}{c}{no pruning} & 0.6541 & 0.6364 \\
    \cmidrule(){1-9}
    \multirow{3}{*}{\makecell{ProtoPShare\\(ours)}} & 1960 & \checkmark & \checkmark & \checkmark & 0.7864 & 0.7767 & \\
     & 1960 & & \checkmark & \checkmark & 0.7902 & 0.7903 \\
     & 1960 & & & \checkmark & \textbf{0.8638} & \textbf{0.8481} \\
    \cmidrule[1pt](){1-9}
    \end{tabular*}
\caption{Accuracy (higher is better) for different models with the same final number of prototypes $|P|$ trained on the Stanford Cars dataset.}
\label{tab:diff_optims}
\end{table}



\end{document}